\documentclass[letterpaper]{article} 
\usepackage{aaai2026}  
\usepackage{times}  
\usepackage{helvet}  
\usepackage{courier}  
\usepackage[hyphens]{url}  
\usepackage{graphicx} 
\usepackage{natbib}  
\setcitestyle{numbers,square}
\usepackage{caption} 
\usepackage{times}  

\usepackage{helvet}  
\usepackage{courier}  
\usepackage[hyphens]{url}  
\usepackage{graphicx} 
\usepackage{algorithm}
\usepackage{colortbl}
\usepackage{algorithmic}
\usepackage{booktabs}
\usepackage{amsmath}
\usepackage{caption} 
\usepackage{cuted}
\usepackage{color,xcolor}
\usepackage{algorithm}
\usepackage{algorithmic}
\usepackage{newfloat}
\usepackage{listings}
\usepackage{algorithm}
\usepackage{algorithmic}

\usepackage{newfloat}
\usepackage{listings}
\DeclareCaptionStyle{ruled}{labelfont=normalfont,labelsep=colon,strut=off} 
\floatstyle{ruled}
\newfloat{listing}{tb}{lst}{}
\floatname{listing}{Listing}
\title{FEB-Cache: Frequency‑Guided Exposure Bias Reduction for Enhancing Diffusion Transformer Caching}
\author{
    Zhen Zou,
    Feng Zhao
}
\affiliations{
    \textsuperscript{\rm 1}MoE Key Laboratory of Brain-inspired Intelligent Perception and Cognition, University of Science and Technology of China\\

    zouzhen@mail.ustc.edu.cn
}

\usepackage{bibentry}

\begin{document}

\maketitle

\begin{abstract}
Diffusion Transformer (DiT) has exhibited impressive generation capabilities but faces great challenges due to its high  computational complexity.    To address this issue, various methods, notably feature caching, have been introduced.
However,  these approaches focus on aligning non-cache diffusion without analyzing why caching damage the generation processes.  In this paper, we first confirm that the cache greatly amplifies the exposure bias, resulting in a decline in the generation quality. 
However, directly applying noise scaling is challenging for this issue due to the non-smoothness of exposure bias.
We found that this phenomenon stems from the mismatch between its frequency response characteristics and the simple cache of Attention and MLP. 
Since these two components exhibit unique preferences for frequency signals, which provides us with a caching strategy to separate Attention and MLP to achieve an enhanced fit of exposure bias and reduce it. 
Based on this, we introduced FEB-Cache,  a joint caching strategy that aligns with the non-exposed bias diffusion process (which gives us a higher performance cap)  of caching Attention and MLP based on the frequency-guided cache table.  
Our approach combines a comprehensive understanding of the caching mechanism and offers a new perspective on leveraging caching to accelerate the diffusion process.
Empirical results indicate that FEB-Cache optimizes model performance while concurrently facilitating acceleration.  Code is available at \url{https://github.com/aSleepyTree/EB-Cache}.
\end{abstract}
\section{Introduction}
\label{sec:intro}

The advent of Diffusion Transformer (DiT) \cite{peebles2023scalable} has catalyzed the generative modeling field, particularly in image \cite{goodfellow2020generative,ho2020denoising,rombach2022high} and video generation \cite{brooks2024video}. DiT has showcased an unprecedented ability to synthesize high-fidelity visual content, offering a powerful alternative to previous U-Net architecture diffusion models \cite{ho2020denoising,dhariwal2021diffusion}. However, the remarkable capabilities of DiT are accompanied by a significant computational cost, which becomes increasingly burdensome when tackling large amounts of samples \cite{vaswani2017attention}. 
As a result, the computational cost escalates dramatically with even modest increases in resolution, posing a substantial challenge to the practical deployment of DiTs  \cite{chen2023pixart,chen2024pixart_s,lu2024fit}.

\begin{figure}[tp]
    \centering
    \includegraphics[width=1\linewidth]{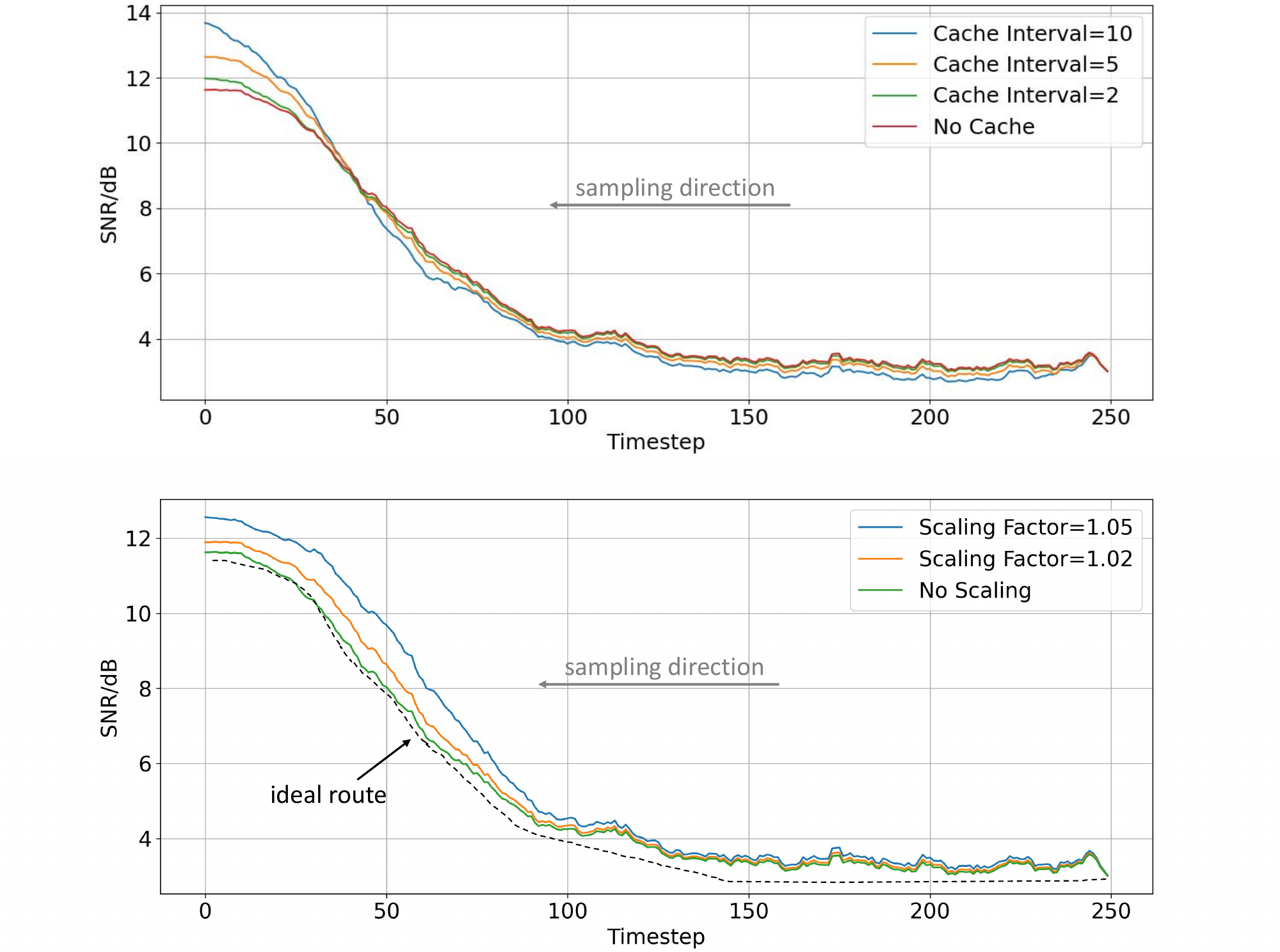}
    \captionsetup{font={normalsize}}
    \caption{(a) Caching increases the SNR of images.
    (b) For ease of observation, we use factors larger than 1 to amplify exposure bias, which increases the SNR of the images.
    }
    \label{fig1:snr_cache_and_eb}
\end{figure}

Feature caching \cite{wimbauer2024cache,ma2024deepcache,selvaraju2024fora,chen2024delta} is a common approach to address this challenge. As a pioneer, DeepCache \cite{ma2024deepcache} avoids deep-level calculations by caching shallow features in U-Net-based diffusion models. Recently, FORA\cite{selvaraju2024fora} and Learning-to-Cache (L2C) \cite{ma2024learning} adapt a caching strategy into DiT, yielding commendable results.  While feature caching has provided a promising pathway to expedite generation, it compromises the quality of the generated images at times. 
These methods concentrate on aligning cached diffusion with the steps of standard diffusion, and they have not explored the damage of caching on visual effects. 
\label{use_fig2}
As shown in Fig.~\ref{fig1:snr_cache_and_eb}, our analysis begins with the Signal-to-Noise Ratio (SNR) curve of the generation process. An interesting phenomenon is that the SNR of the cache-enhanced image is improved, especially in the stage with high SNR. This reminds us to explore the relationship between cache and the potential effects that  enhance the SNR of images. 
Exposure bias, an inherent bias in diffusion models, amplifies the SNR of an image by reducing more noise \cite{li2023alleviating,ningelucidating,ning2023input}. We prove that exposure bias is positively correlated with the variance of prediction error for the final generated results (in Appendix), and cache amplifies this error variance, leading to the amplification of exposure bias and subsequently a decline in generation quality.

However, the SNR curves between cache and exposure bias in Fig.~\ref{fig1:snr_cache_and_eb} do not exactly match. Exposure bias amplifies the SNR earlier than cache, which indicates that the intensification of exposure bias by cache is not a simple amplification effect. Considering the selectivity of SNR for different frequencies of the signal, we believe that this is caused by the mismatch of the frequency signal selection mechanism between vanilla cache and exposure bias. 
We find that the influence of exposure bias on low-frequency mainly occurs in the generation's early stage, while its impact on high-frequency mainly occurs in the later stage. However, vanilla cache ignores such a preference for different stages and thus struggles to reduce the exposure bias well.

\begin{figure*}[tp]
    \centering
    
    \includegraphics[width=1 \linewidth]{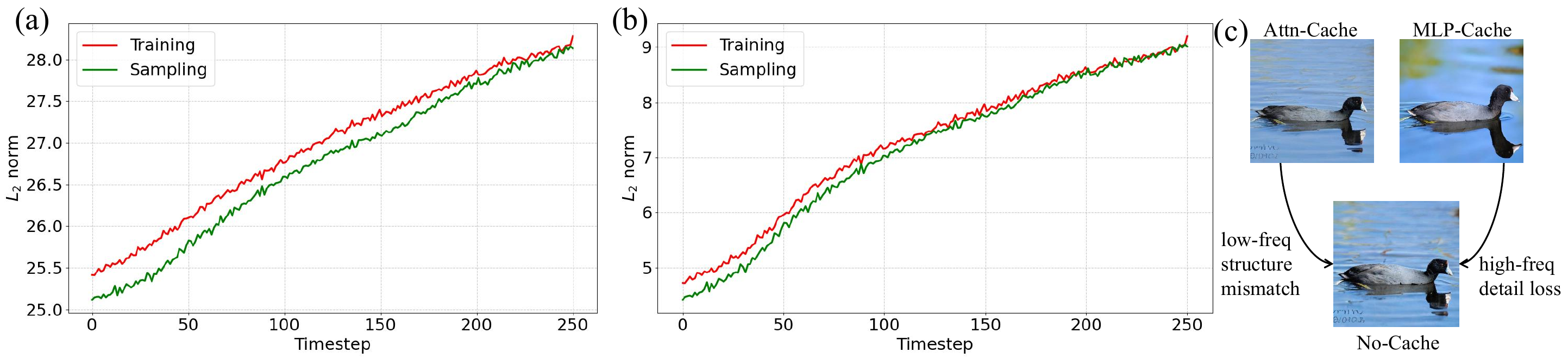}
    \captionsetup{font={normalsize}}
    \caption{(a) $L_2$ norm of low-frequency component for intermediate noisy images.
    (b) $L_2$ norm of high-frequency component for intermediate noisy images.  Note that calculations are performed in image space decoded by VAE \cite{kingma2013auto}. 
    (c) Images generated under Attn or MLP-Cache show different frequency influence. 
    }
    \label{fig2:main}
\end{figure*}

Previous studies \cite{bai2022improving, park2022vision, wang2022vtc} have found that the basic components of transformer, Attention(Attn) and MLP, possess low and high frequency preferences, respectively. This prompts us to explore whether this preference still exists in the feature cache of DiT. As shown in Fig.~\ref{fig2:main}, we cached Attn and MLP respectively and found that caching only MLP would cause the image to lack high-frequency details, while caching only Attn would lead to the destruction of the low-frequency structural information of the image. This preference, coupled with the frequency preference of exposure bias, provides us with a way to better reduce exposure bias with noise scaling that is often employed in exposure bias reduction.

In this paper, we introduce a new cache mechanism to cache Attn and MLP based on their preference for enlarging different frequencies in exposure bias.
Specifically, we propose Frequency-guided Exposure Bias Reduction Cache (FEB-Cache) to flexibly manage exposure bias for caching Attn and MLP in different stages.
We design the FEB-Cache based on the error introduced by uniform caching, which ignores the frequency preference of exposure bias.
Our method is based on alignment with non-exposure bias diffusion, where caching plays a dual role in acceleration and balancing exposure bias, while scaling noise throughout the diffusion process to reduce it.

\begin{itemize} 
    \item[$\bullet$] We demonstrate that caching leads to a decline in generation quality by amplifying exposure bias.
    
    \item[$\bullet$] By analyzing the frequency preference of exposure bias and the frequency division of labor in the generation process of Attn and MLP, we propose FEB-Cache to reduce exposure bias and accelerate generation.

    \item[$\bullet$] Extensive experiments demonstrate the effectiveness and reliability of our method in different scenarios.
\end{itemize}

\section{Related Work}

\label{sec:related}

\subsection{Acceleration of Diffusion Transformers}

Diffusion model's huge computational burden comes from its iterative denoising during the inference process, which can't be performed in parallel due to the Markov chain \cite{ho2020denoising}. A variety of acceleration techniques have been proposed to overcome this challenge \cite{bolya2023token,bolya2022token,lou2024token,yuan2024ditfastattn,Structuralpruning,zhang2024laptop,shang2023post,so2024temporal}. Some methods rely on model compression or quantization to reduce the model size or computational accuracy \cite{dong2024ditas,kim2023architectural,salimans2022progressive,luo2023latent,shang2023post,so2024temporal}. In addition, some methods explore efficient solvers and strategies to minimize the steps required to generate high-quality images \cite{lu2022dpm,zhang2022fast,karras2022elucidating,lu2022dpm+,liu2023oms,song2020denoising}. 

Another way is to reduce the average amount of calculation per step \cite{yang2023diffusion,li2023faster}. Some works \cite{selvaraju2024fora,ma2024learning,chen2024delta} introduce the idea of feature caching and adapt it to the transformer architecture. 
FORA \cite{selvaraju2024fora} replaces layers with the output of previous step. $\Delta$-DiT \cite{chen2024delta} skips layers by caching feature residuals instead of feature maps. By reusing similar features, they significantly reduce the computational cost.

\subsection{Exposure Bias in Diffusion Models}
Exposure bias in Diffusion Models is a significant issue that has attracted considerable research attention in the field of generative modeling \cite{li2023alleviating,ningelucidating,ning2023input,tsonis2023mitigating}.  Exposure bias refers to the discrepancy between the training and inference processes of a model, which can lead to suboptimal performance during generation.  In the context of DPMs, this bias is exacerbated by the fact that the models typically require numerous iterative steps to synthesize high-quality images, making the models prone to accumulate errors to an explosion during the inference process.

Some methods have been proposed to solve this problem. \citet{li2023alleviating} believe that although there is a deviation between the prediction of $\hat{x}^t_0$ and the ground truth $x_0$, there may be another $\hat{x}_0^{t_s}$ better coupled with $x_0$. The authors propose a Time-Shift Sampler to search for such $t_s$ around $t$ to avoid exposure bias. \citet{ningelucidating} find that exposure bias makes the network predict \textit{slightly too much noise} and thus impairs the network performance. The authors propose a method named Epsilon Scaling to weaken this effect by scaling the prediction noise of the network. However, since the influence of exposure bias is uneven throughout the diffusion process, determining hyperparameters remains a problem. {In this paper, we modulate exposure bias with separated feature caching, achieving a uniform exposure bias throughout the diffusion process, which can be dealt with a constant scaling factor.}

\label{Exposure Bias in Feature Caching}

\section{Motivation}
\label{sec:Motivation}

\subsection{Exposure Bias in Feature Caching}

Exposure bias comes from the inconsistency between training and inference in diffusion models \cite{bao2022analytic}. 
\begin{footnotesize}
\begin{equation}
\begin{aligned}
    \label{eq: temp}
    \hat{x}_0^{t-1} &= x_0 + e_{t-1}\epsilon_{t-1} \quad (\epsilon_{t-1} \sim \mathcal{N}(0, I)) ,   \\
    \hat{x}_{t-1} &= \sqrt{\bar{\alpha}_{t-1}}x_0 + \sqrt{1 - \bar{\alpha}_{t-1} + \underbrace{\left(\frac{\sqrt{\bar{\alpha}_{t-1}}\beta_t}{1-\bar{\alpha}_t}e_t\right)^2}_{\text{Exposure Bias Term}}} \epsilon_1 ,
\end{aligned}
\end{equation}
\end{footnotesize}
\noindent where $e_t$ donates the variance between $\hat{x}^{t-1}_0$ (predicted $x_0$) and real $x_0$ at step $t-1$. Feature caching reuses intermediate features in subsequent steps. The exposure bias term is positively related to the variance term (See Appendix). Under the assumption that errors in adjacent steps are correlated with a coefficient $\rho\ (0 \leq \rho \leq 1)$ and for N cached steps, the total variance of accumulated error is:
\begin{equation}
\label{eq: super linear}
\text{Var}\left(\sum_{k=1}^N \epsilon_{t-k}\right) = N + 2\sum_{d=1}^{N-1}(N-d)\rho^d > N,
\end{equation}
where the covariance term dominates when $\rho \rightarrow 1$.

A common approach to mitigate exposure bias is noise scaling, which adjusts predicted noise to counteract accumulated errors. However, this strategy struggles with the non-uniform nature of exposure bias: it exhibits non-uniform characteristics across different frequency components, varying dynamically across the diffusion process. It makes static scaling prone to over-correction in some stages and under-correction in others.
Our key insight is that if exposure bias can be fitted and modulated to a uniform state, which is  consistent across the diffusion process—noise scaling could effectively reduce it. But as shown in Fig.~\ref{fig1:snr_cache_and_eb}, vanilla cache can not perfectly fit exposure bias. In Fig.~\ref{fig2:main}(a) and 2(b), we decompose each noisy image generated within the training process into high and low-frequency components with FFT and calculate its $L_2$ norm for comparison with the training process. They tell us that  exposure bias has a selective preference for different frequency components of the data, which vanilla cache mechanism does not account. For low-frequency components, exposure bias begins to exert its impact early in the denoising process when the noise level is still high. In contrast, high-frequency components are affected by exposure bias only in the later stages of denoising when the noise has been significantly reduced.

\subsection{Modulate Exposure Bias with Frequency-guided Feature Caching}
\label{Modulate Exposure Bias with Feature Caching}

The key to achieving perfect fitting lies in leveraging the frequency selectivity of different components within the transformer model.
Previous studies \cite{bai2022improving, park2022vision, wang2022vtc}  have highlighted that the self-attention mechanism within Transformers predominantly handles low-frequency information, while MLP processes high-frequency information more. This functional dichotomy persists in the feature caching mechanism of DiT. Fig.~\ref{fig2:main}(c) reveals that caching Attn and MLP components of the transformer show distinct effects on the image signal.  Specifically, caching Attn primarily influences the low-frequency structures of the image, such as the overall contour and large-scale features, while the MLP caching mainly affects the high-frequency details, including texture and fine-grained information.

This frequency selectivity of different transformer components provides us with a valuable insight: by separating caching for MLP and Attn, we can independently modulate their impacts on low-frequency and high-frequency signal.  This separation allows the cache to perfectly fit the non-uniform exposure bias, as we can tailor the cache strategies for each component to match the specific characteristics of exposure bias on corresponding frequency components.

\begin{figure*}[tp]
    \centering
    \includegraphics[width=1 \linewidth]{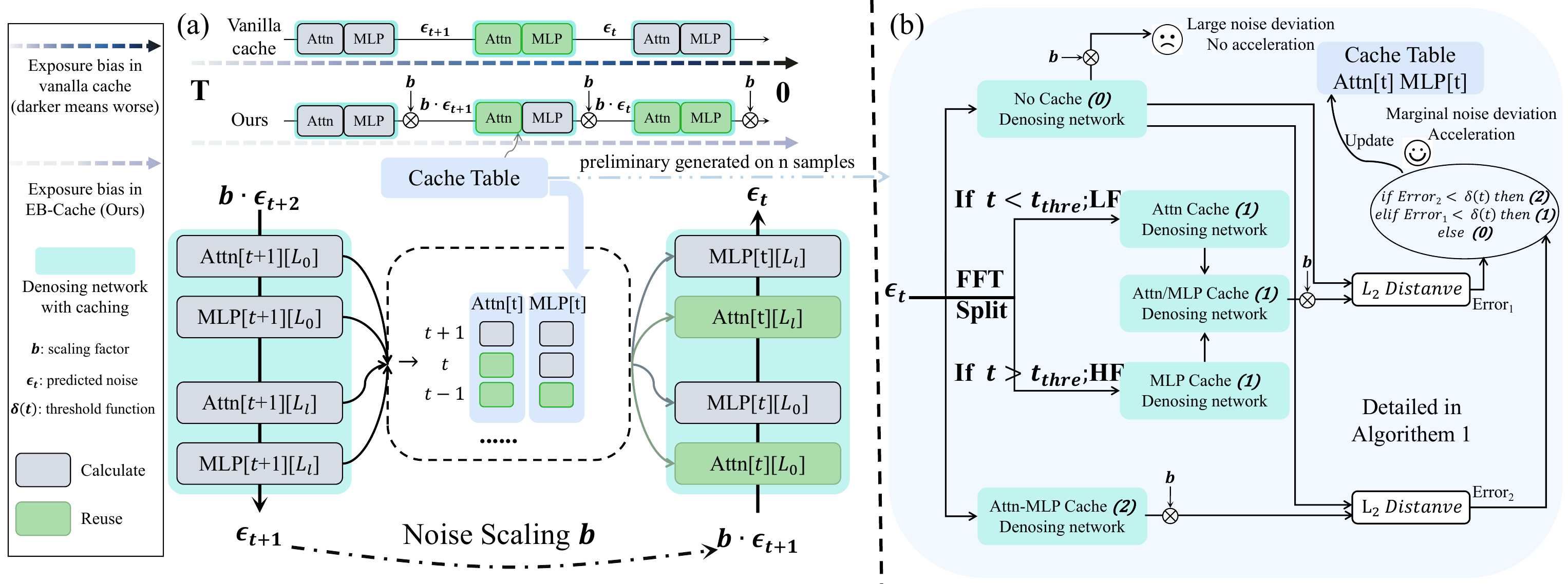}
    \captionsetup{font={normalsize}}
    \caption{(s)Illustration of FEB-Cache. 
    FEB-Cache reduces exposure bias while achieving laudable speed up. Frequency-guided Cache Table is pre-generated as shown in (b) on $n$ samples. Details can be found in Algorithm.\ref{algorithm}. 
    }
    \label{fig3:main}
\end{figure*}
\section{Method}
\label{sec:Method}

\begin{table*}[t]
    \centering
    \captionsetup{font={normalsize,stretch=1.25},justification=raggedright}
    
    \resizebox{0.9\linewidth}{!}{
    \begin{tabular}{l|c|ccc|ccccc}
        
        \toprule
        Methods & NFE  & FLOPs(T)& Latency(s) & Speedup &  FID$\downarrow$ & sFID$\downarrow$ &IS$\uparrow$ & Precision$\uparrow$ & Recall$\uparrow$  \\
        \midrule
        \multicolumn{9}{c}{DiT-XL/2 (ImageNet 256×256) (cfg=1.5)} \\
        
        \midrule 
        
        DiT \cite{peebles2023scalable} & 250 & 29.67&77.42 & 1.00 & 2.27 & 4.62 & 279.36 & 0.83 & 0.57 \\
        
        
        DiT \cite{peebles2023scalable} & 165 & 19.58& 51.36 & 1.51 & 2.36 & 5.11 & 265.71 & 0.81 & 0.57 \\
        
        FORA(N=2) \cite{selvaraju2024fora} & 250& 16.27  & 49.96 & 1.55 & 2.41 & 5.38 & 266.81	& 0.81 & 0.59 \\
        
        DiTFastAttn \cite{yuan2024ditfastattn} & 250& 16.40  & 50.90 & 1.52 & 2.52 & 5.03 & 269.58	& 0.82 & 0.58 \\
        
        L2C \cite{ma2024learning} & 250&20.18  & 57.35 & 1.35 & 2.31	& \textbf{4.90} & 271.97 & 0.81 & 0.58 \\
        
        \textbf{Ours} & 250&16.53  & 50.52 & 1.53 & \textbf{2.27} & 4.90 & \textbf{273.31} & \textbf{0.82} & \textbf{0.59} \\
        
        DiT \cite{peebles2023scalable} & 100&11.93 & 30.96 & 2.50 & 2.57 & 5.73& 251.69  & 0.81 & 0.57 \\
        
        FORA(N=4) \cite{selvaraju2024fora} & 250&11.31 & 31.49 & 2.46 & 3.46	& 8.06 & 243.17	& 0.78 & \textbf{0.59} \\
        
        DiTFastAttn \cite{yuan2024ditfastattn}  & 250 &11.40 & 30.21 & 2.51 & 3.06	& 7.43 & 253.09 & 0.79 & 0.56 \\
        
        ToCa(N=4,R=90\%) \cite{zou2024accelerating}& 250&11.87  & 31.79 & 2.44 & 2.52	& 5.40 & 260.19 & 0.81 & 0.58 \\
        
        ToCa + Ours & 250&11.36  & 30.90 & 2.51 & 2.43	& 5.29 & 260.97 & 0.81 & 0.58 \\
        
        \textbf{Ours} & 250 &11.73& 31.27 & 2.48 & \textbf{2.40} & \textbf{5.16} & \textbf{263.97}	& \textbf{0.81} & 0.58 \\

        \midrule
        
        DiT  \cite{peebles2023scalable} & 50 & 5.98&15.52 & 1.00 & 3.69 & 8.56 & 220.05 & 0.78 & 0.57 \\
        
        
        DiT  \cite{peebles2023scalable} & 33&3.87 & 10.37 & 1.50 & 4.65 & 8.56 & 197.02 & 0.78 & 0.57 \\
        
        FORA(N=2) \cite{selvaraju2024fora} & 50 &3.51 & 10.01 & 1.55 & 8.45	& 14.28 & 178.93 & 0.80 & 0.56 \\
        
        DiTFastAttn \cite{yuan2024ditfastattn}  & 50 &3.66  & 10.21 & 1.52 & 5.66	& 10.34 & 164.11 & 0.79 & 0.56 \\
        
        L2C \cite{ma2024learning} & 50 &4.03 & 11.91 & 1.30 &3.73	& 8.61 & 209.88 & 0.80 & 0.57 \\
        
        \textbf{Ours} & 50 &3.49 & 10.40 & 1.49 & \textbf{3.05}	& \textbf{5.98} & \textbf{216.94}	& \textbf{0.80} & \textbf{0.57} \\
        
        DiT  \cite{peebles2023scalable} & 20&2.47 & 6.20 & 2.50 & 15.92 & 19.22 & 173.58 & \textbf{0.78} & 0.57 \\
        
        FORA(N=4) \cite{selvaraju2024fora} & 50&2.39 & 6.34 & 2.45 & 88.21	& 102.40 & 100.03	& 0.51 & 0.46 \\
        
        DiTFastAttn \cite{yuan2024ditfastattn}  & 50 &2.43  & 6.33 & 2.45 & 61.07	& 70.05 & 101.74 & 0.61 & 0.52 \\
        
        ToCa(N=4,R=90\%) \cite{zou2024accelerating}& 50&2.51 & 6.47 & 2.40 & 13.77	& 20.10 & 177.60	& 0.73 & 0.57 \\
        
        ToCa + Ours& 50&2.43 & 6.28 & 2.47 & 13.10	& 18.52 & 181.72	& 0.73 & 0.57 \\
        
        \textbf{Ours} & 50&2.40 & 6.30 & 2.46 & \textbf{10.32}	& \textbf{15.82} & \textbf{195.61}	& 0.76 & \textbf{0.58} \\

        \midrule
        \multicolumn{9}{c}{DiT-XL/2 (ImageNet 512×512) (cfg=1.5)} \\
        \midrule
        
        DiT  \cite{peebles2023scalable} & 100&11.87 & 88.73 & 1.00 & 5.83 & 17.30 & 213.57 & 0.83 & 0.63 \\

        DiT  \cite{peebles2023scalable} & 60&7.28 & 53.75 & 1.65 & 6.10 & 17.21 & 206.15 & 0.83 & 0.62 \\
        
        FORA(N=2) \cite{selvaraju2024fora} & 100 &7.33 & 53.28 & 1.67 &  6.36	& 19.20 & 202.90	& 0.84 & 0.65 \\
        
        DiTFastAttn \cite{yuan2024ditfastattn}  & 100 &7.27  & 53.88 & 1.65 & 6.10	& 19.13 & 208.80 & 0.83 & 0.64 \\
        
        L2C \cite{ma2024learning} & 100&7.15  & 54.62 & 1.63 & 6.01	& 17.56 & 208.38 & 0.84 & 0.64 \\
        
        ToCa(N=3,R=70\%) \cite{zou2024accelerating}& 100&7.40 & 54.93 & 1.62 & 6.04	& 17.29 & 206.90	& 0.83 & 0.63 \\
        
        ToCa + Ours& 100& 7.17 & 53.29 & 1.67 & 5.97& 17.11 & 208.99	& 0.83 & 0.65 \\
        
        \textbf{Ours} & 100 &7.20 & 52.37 & 1.69 & \textbf{5.91}	& \textbf{16.81} & \textbf{209.17} & \textbf{0.84} & \textbf{0.65} \\

    \bottomrule
    
    \end{tabular}
    
    }
    \caption{Comparison on ImageNet with DiT-XL/2 DDPM. Note that L2C degenerates to FORA when speedup ratio $> 2.$ } 
    \label{tbl:main_table}
\end{table*}

\subsection{Overall Framework}
This section details the overall framework and the design of the method.
As shows in Fig.~\ref{fig3:main}, our method combines noise scaling and a separated cache table to modulate exposure bias effectively. These two operations work in tandem to counteract the non-uniform amplification of exposure bias while preserving the computational efficiency of cache mechanisms.
In Appendix, we show more details and optimization of the signal in our method.

\noindent \textbf{Adaptive Noise Scaling:}
We introduce a noise scaling factor to adjust the magnitude of noise predicted by the network, thereby mitigating the cumulative effects of exposure bias.  The scaling factor is dynamically adjusted across the diffusion process using stage-specific functions to account for the time-varying nature of exposure bias, with distinct formulations for high-noise and low-noise phases. In the early stages of diffusion (high noise levels, corresponding to denoising steps where $t \in (t_{thre},1]$), we employ a scaling factor based on a base value like $b_h = 0.98$. In the later stages (low noise levels, where $t \in [0,t_{thre}]$), we use a scaling factor with a base value like $b_l=0.96$: 

\begin{equation}
b(t)  = 
\begin{cases}
    b_h+(1-b_h) \cdot e^{\frac{-5(1-t)}{1-t_{thre}}} & t_{thre} \leq t \leq T, \\
    b_l+(b_h-b_l) \cdot e^{\frac{5(t-t_{thre})}{1-t_{thre}}} & 0 \leq t < t_{thre},
\end{cases}
\end{equation}
where $t$ represents the denoising step progressing from $T$ (full noise) to $0$ (clean signal), with $t_{thre}=0.4 \cdot T$ serving as the threshold separating the two stages. This formulation ensures the scaling factor evolves smoothly within each phase: during early stages, when exposure bias has not yet accumulated significantly, the scaling remains close to $b_h$ to provide mild but adaptive modulation.  In later stages, as exposure bias effects compound over iterations, the scaling shifts toward $b_l$ to deliver stronger, targeted modulation that prevents error amplification.

\noindent \textbf{Separated Cache Table}
To complement the noise scaling, we design a separated cache table that maintains distinct caches for Attn and MLP components.  This separation is motivated by the observation that Attn and MLP affect low-frequency and high-frequency components, respectively (Fig. 2(c)).  By isolating these caches, we enable targeted modulation of exposure bias across different frequency bands, ensuring the cache mechanism can "fit" the non-uniform characteristics of exposure bias as discussed.

\subsection{Frequency-Guided Separated Cache Table}
The core of our approach lies in tailoring the cache table to the frequency-dependent behavior of exposure bias. As observed in Fig. 2(a) and 2(b), exposure bias impacts low-frequency components early in denoising and high-frequency components later. To align with this pattern, we dynamically adjust the cache table to prioritize different components (MLP or Attn) at different stages, using a dynamic selection of cache candidates.

\noindent \textbf{Early Stages}: During high-noise phases, exposure bias primarily impacts low-frequency components, while high-frequency details remain relatively less affected.   Since MLP components are responsible for processing high-frequency signals (Fig. 2(c)), caching MLP outputs becomes advantageous in this stage.   Caching MLP outputs allows us to retain and reuse stable high-frequency information without exacerbating the already vulnerable low-frequency components. Thus, the cache candidates in early stages are [No Cache, MLP Cache, MLP-Attn Cache], ensuring that high-frequency processing remains robust while preventing further amplification of low-frequency exposure bias.

\noindent \textbf{Late Stages}: As denoising progresses and noise decreases, exposure bias shifts to predominantly affect high-frequency components, with low-frequency structures becoming more stable.  In this phase, attention components— which govern low-frequency structures—are prioritized for caching.  Caching Attn outputs helps preserve the stability of low-frequency features, which are now less susceptible to exposure bias.  Conversely, MLP components (handling high-frequency processing) are not prioritized for standalone caching here, as this could amplify exposure bias on the now-vulnerable high-frequency details.  The cache candidates are therefore adjusted to [No Cache, Attn Cache, MLP-Attn Cache], enabling targeted reuse of Attn outputs to maintain low-frequency stability while allowing controlled use of MLP caching only when combined with Attn.

To implement the stage-aware frequency-selective caching strategy, Algorithm.~\ref{algorithm} operationalizes the candidate selection through a greedy approach that evaluates error metrics to determine the optimal cache state at each denoising step.  The cache table maintains three candidate states for each stage: No Cache (no historical information reused), MLP or Attn Cache (only MLP or Attn outputs cached), and MLP-Attn Cache (both MLP and Attn outputs cached).

For each step $t$, it computes three error metrics: the baseline error $E_{ori}(t)$ from the original network prediction, the error $E_{Attn-MLP}(t)$ when both attention and MLP outputs are cached (scaled by the noise scaling factor $b(t)$, and the error $E_{Candidate}(t)$ for the stage-specific candidate (MLP in early stages, Attn in late stages). By comparing these errors against a time-varying threshold function $\delta(t)$, the algorithm tags each step with the appropriate cache state. 
Through this data-driven selection, the cache table dynamically adapts to the evolving noise levels and exposure bias characteristics, reinforcing the synergy between noise scaling and frequency-guided caching to achieve uniform modulation of exposure bias.

\begin{algorithm}[h]
   \caption{Frequency-guided Cache Table Generation}
   \label{algorithm}
\begin{algorithmic}[1]
        
        \STATE \textbf{Input:} Denoising Network ${\epsilon_{\theta}}$, Scaling Factor $b(t)$,  Total Step $T$, Error Threshold Function $\delta (t)$
        \STATE Initialize Cache Table $\gets \emptyset$
        \REPEAT
        \STATE $x_T \sim \mathcal{N}({0}, {I})$
        \STATE $x_{T-1} = \frac{1}{\sqrt{\alpha_T}}(x_T - \frac{\beta_T}{\sqrt{1-\bar{\alpha}_T}} {\epsilon_{\theta}}(x_T, T) \cdot b)$
        \FOR {$t := T-2, ..., 1$}

        \STATE $Candidate \gets MLP~or~Attn$
        \STATE $E_{ori}(t) \gets \Vert{\epsilon_{\theta}}(x_t, t)\Vert_2$    

        \STATE  $E_{Attn-MLP}(t) \gets \Vert{b(t) \cdot \epsilon^{'}_{\theta}}(x_t, t)\Vert_2 $    

        \STATE  $E_{Candidate}(t) \gets \Vert{b(t) \cdot \epsilon^{''}_{\theta}}(x_t, t)\Vert_2 $

         \IF{ $  \Vert E_{ori}(t) - E_{Attn-MLP}(t) \Vert_1  < \delta (t)$}
        
        \STATE Tag $t$ :\textsc{ Attn-MLP Cache }

         \ELSIF{   $\Vert E_{ori}(t) - E_{Candidate}(t) \Vert_1  < \delta (t)$} \STATE Tag $t$ : \textsc{Candidate Cache }
        \ELSE \STATE     Tag $t$ :\textsc{ No Cache}
        \ENDIF
        \ENDFOR
        \UNTIL {$n$ samples}
        \STATE Weight Cache Table
    \end{algorithmic}
\end{algorithm}

\section{Experiment}

This section presents an overview of our experimental setup, including datasets, evaluation models, experimental configurations, assessment metrics used and results.

\begin{figure*}[tp]
    \centering
    \includegraphics[width=\linewidth]{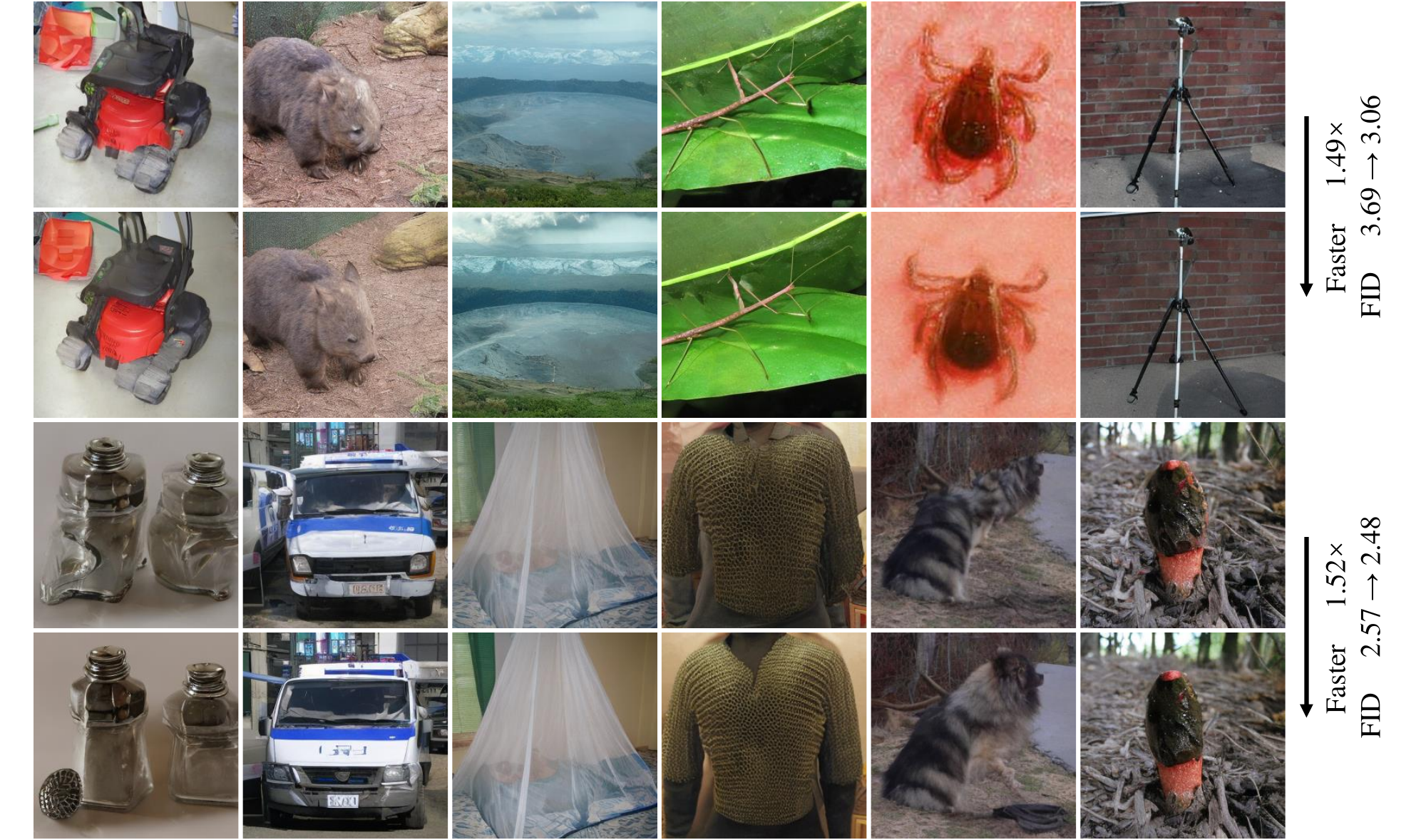}
    \captionsetup{font={normalsize}}
    \caption{Qualitative comparison on ImageNet 256$\times$256 with 50 and 100 NFEs.}  
    \label{fig: visualization}
\end{figure*}

\begin{table}[tbh]
    \centering
    \resizebox{0.47 \textwidth}{!}{
    \begin{tabular}{l|cccccc}
        \hline
         Methods &  FLOPs(T)&Latency(s) & FID $\downarrow$ &  sFID $\downarrow$ & IS $\uparrow$
        \\
        \hline
        DiT (T=20)  \cite{peebles2023scalable} &  2.37& 5.93  & 3.48 &  4.88 & 222.97
        \\
        DiT (T=16)  \cite{peebles2023scalable} & 1.90 &4.63& 4.66  &  5.71 & 209.72
        \\
        ToCa (N=2, R=40\%) \cite{zou2024accelerating}  & 1.88 &4.73& 4.17  &  5.13 & 213.01
        \\
        DiTFastAttn \cite{yuan2024ditfastattn} & {1.88}&4.67  & 4.09  &  5.21 & 219.30
        \\
        $\Delta$-DiT \cite{chen2024delta} & 1.91 &{4.61} & 3.97  &  5.19 & 211.62
        \\
        L2C \cite{ma2024learning} &{1.86} &4.69 & {3.46}  &  {4.66} & \textbf{226.76}
        \\
        Ours  & {1.88}&{4.61}  & \textbf{3.37}  &  \textbf{4.65} & {225.19}
        \\
        \hline
    \end{tabular}}
    \caption{256 $\times$ 256 ImageNet DiT-XL/2 with DDIM.}
    \label{tab: main_table_ddim}
\end{table}

\begin{table}[tbh]
    \centering
    \resizebox{0.47 \textwidth}{!}{
    \begin{tabular}{l|ccccc}
        
        \hline
        \multicolumn{5}{c}{1024 $\times$ 1024 PixArt-$\Sigma$ DPM-Solver} \\
        \hline
         Methods &  FLOPs(T)&Latency(s) & FID $\downarrow$ &  CLIP $\uparrow$ & IS $\uparrow$
        \\
        \hline
        PixArt-$\Sigma$ (T=20) \cite{chen2024pixart_s} &  173.08 &1.39  & 83.50 &  32.01 & 23.61
        \\
        PixArt-$\Sigma$ (T=13) \cite{chen2024pixart_s} & 112.52 &0.93 & 84.97  &  31.84 & 23.10
        \\
        ToCa (N = 3, R = 50\%) \cite{zou2024accelerating} & {109.93} & {0.95} & 83.07 & {32.10} & 23.41
        \\
        DiTFastAttn \cite{yuan2024ditfastattn} & 113.00 &0.97 & {82.06}  & 31.99 & 23.17
        \\
        $\Delta$-DiT \cite{chen2024delta} & {112.25}&{0.95}  & 82.72  &  32.03 & {23.67}
        \\
        Ours  & 113.52 &0.97  & \textbf{81.14} &  \textbf{32.37} & \textbf{25.36}
        \\
        \hline
        \multicolumn{5}{c}{2048 $\times$ 2048 PixArt-$\Sigma$ DPM-Solver} \\
        \hline
        PixArt-$\Sigma$ (T=20) \cite{chen2024pixart_s} &  693.88&5.39  & 139.79 &  37.81 & 21.07
        \\
        PixArt-$\Sigma$ (T=13) \cite{chen2024pixart_s} & 450.01&3.67  & 143.69  &  37.19 & 20.49
        \\
        ToCa (N = 3, R = 50\%) \cite{zou2024accelerating} & {428.93}&{3.62} & 140.77 & 37.57 & {21.05}
        \\
        DiTFastAttn \cite{yuan2024ditfastattn} & 451.33 &3.79 & \textbf{138.67}  &  {37.69} & 21.01
        \\
        $\Delta$-DiT \cite{chen2024delta} & {446.29}&3.78  & 141.80  &  37.66 & 20.94
        \\
        Ours  & 449.37&{3.73}  & {140.03}  &  \textbf{38.30} & \textbf{21.29}
        \\
        \hline
    \end{tabular}}
    \caption{PixArt-$\Sigma$ T2I generation.}
    \label{tab: main_table_pixart}
\end{table}

\begin{table}[tbh]
    \centering
    \resizebox{0.47 \textwidth}{!}{
    \begin{tabular}{l|cccccc}

        \hline
         Methods &  FLOPs(T)&Latency(s) & VBench(\%) $\uparrow$ &  PSNR $\uparrow$ & SSIM $\uparrow$& LPIPS $\downarrow$
        \\
        \hline
        CogVideoX (T=50) \cite{yang2024cogvideox} & 8417.40 &142.10&80.91&-&-&-
        \\
        CogVideoX (T=35) \cite{yang2024cogvideox} & 5892.18 &98.73&78.19&25.70&0.9183&0.0541
        \\
        PAB$_{357}$ \cite{zhao2024real} &{5905.38}&{97.52}&{78.66}&25.48&{0.9213}&{0.0529}
        \\
        $\Delta$-DiT \cite{chen2024delta} & {5861.42}&{96.61}&78.21&{25.51}&0.9207&0.0537
        \\
        Ours  & 5910.74&98.01&\textbf{78.79}&\textbf{25.81}&\textbf{0.9219}&\textbf{0.0521}
        \\
        \hline
    \end{tabular}}
    \caption{720 $\times$ 480 CogVideoX-2B T2V generation.}
    \label{tab: main_table_video}
\end{table}

\begin{table}[tbh]
    \centering
    \resizebox{0.45 \textwidth}{!}{
    \begin{tabular}{c|cccccc}
        
        \hline
         &  Latency(s)  & FID $\downarrow$ &  sFID $\downarrow$ & IS $\uparrow$ & Precision$\uparrow$ & Recall$\uparrow$\\
        \hline
        Baseline(T=50)&  15.52  & 3.69 &  8.56 & 220.05 & 0.78 & 0.57
        \\
        \hline
        + Scaling  & 15.48  & 3.61  &  8.61 & 211.79 & 0.78 & 0.57
        \\
        \hline
        + Caching Table & 10.26  & 4.12  &  9.06 & 206.10 & 0.76 & 0.56
        \\
        \hline
        Ours  & 10.31  & \textbf{3.05}  &  \textbf{5.98} & \textbf{216.94} & \textbf{0.80} & \textbf{0.57}
        \\
        \hline
    \end{tabular}}
    \caption{Ablation on method components.}
    \label{tab: ablation_to3}
\end{table}

\begin{table}[tbh]
    \centering
    \resizebox{0.47 \textwidth}{!}{
    \begin{tabular}{c|c|ccccc}
        \hline
         $b_l / b_h$ &baseline&  0.94/0.97 & 0.95/0.975 &  0.96/0.98(Ours) & 0.97/0.985 &  0.98/0.99  
        \\
        \hline
        FID $\downarrow$ &4.65&  3.35  & 3.27 &  3.05 & 3.21& 3.42
        \\
        \hline
        IS $\uparrow$ &197.02& 213.90  & 215.14  &  216.94 & 214.99 & 213.73
        \\
        \hline
    \end{tabular}}
    \caption{Ablation on scaling factors.}
    \label{tab: scaling_factor_ablation}
\end{table}

\begin{table}[tbh]
    \centering
    \resizebox{0.47 \textwidth}{!}{
    \begin{tabular}{c|c|ccc}
        \hline
         functions & baseline & Positive Linear(Ours) & Negative Linear &  Uniform 
        \\
        \hline
        FID $\downarrow$ &4.65&  3.05  & 3.34 &  3.13 
        \\
        \hline
        IS $\uparrow$ &197.02& 216.94  & 213.96  &  215.00 
        \\
        \hline
    \end{tabular}}
    \caption{Ablation on type of threshold functions.}
    \label{tab: threshold_ablation}
\end{table}

\begin{table}[tbh]
    \centering
    \resizebox{0.47 \textwidth}{!}{
    \begin{tabular}{c|c|ccc}
        \hline
         (a, b)&baseline &  (0.1, 0.05) & (0.05, 0.15) &  (0, 0.25) 
        \\
        \hline
        FID $\downarrow$ &  4.65  & 3.51 &  3.05 & 3.77
        \\
        \hline
        IS $\uparrow$ & 197.02  & 211.84  &  216.94  & 210.99
        \\
        \hline
    \end{tabular}}
    \caption{Ablation on parameter of threshold functions.}
    \label{tab: threshold_value_ablation}
\end{table}

\begin{table}[tbh]
    \centering
    \resizebox{0.47 \textwidth}{!}{
    \begin{tabular}{c|c|ccc}
        \hline
         n&baseline &  8 & 32 &  1228 
        \\
        \hline
        FID $\downarrow$ &  4.65  & 3.05 &  3.05 & 3.06
        \\
        \hline
        IS $\uparrow$ & 197.02  & 216.94  &  216.11  & 217.39
        \\
        \hline
    \end{tabular}}
    \caption{Ablation on $n$ in Algorithm.~\ref{algorithm}.}
    \label{tab: n_ablation}
\end{table}

\label{sec:Experiment}
\subsection{Experimental Settings and Baselines}
We explore our methods on the most used transformer-based diffusion model: DiT with different NFEs. Following \cite{ma2024learning}, we test DiT-XL/2 (256×256) with 50 and 250 NFEs and DiT-XL/2 (512×512) with 100 NFEs on ImageNet. Settings are consistent with standard DiT. DDIM works well with few steps, so we test it with 20 NFEs in Tab.~\ref{tab: main_table_ddim}.
In order to verify the generalizability of FEB-Cache in more complex models and tasks, we perform experiments with PixArt-$\Sigma$ \cite{chen2024pixart_s} for T2I and CogvideoX \cite{yang2024cogvideox} for T2V.

Baselines include FORA \cite{selvaraju2024fora}, $\Delta$-DiT~\cite{chen2024delta}, L2C \cite{ma2024learning}, DiTFastAttn \cite{yuan2024ditfastattn}, ToCa \cite{selvaraju2024fora}, and PAB \cite{zhao2024real}. 
All the experiments are conducted on NVIDIA-H100 GPUs.
Due to limited space, please refer to Appendix for complete comparisons. 


\subsection{Main Results}

We present the experimental results in Tab.~\ref{tbl:main_table}, \ref{tab: main_table_ddim}, \ref{tab: main_table_pixart} and \ref{tab: main_table_video}. In general, FEB-Cache perfectly achieves a balance between speed and quality. 
For example, in 50 steps of generation on ImageNet 256×256, our approach achieves a $1.49 \times$ acceleration and an FID of 3.05, which is not only better than the unaccelerated DiT but also better than 3.39 of DiT-ES \cite{ningelucidating} (not shown in the table). This is because the flexible cache strategy makes up for the over-correction of noise caused by scaling in some steps. Original DiT-ES uses a fixed scaling factor throughout the generation process, while our greedy cache strategy makes up for this deficiency of Epsilon Scaling in specific steps by exploiting the amplification effect of cache on exposure bias. We show some samples in Fig.~\ref{fig: visualization}. Refer to Appendix for more.

\begin{figure}[tp]
    \centering
    \includegraphics[width=0.95 \linewidth]{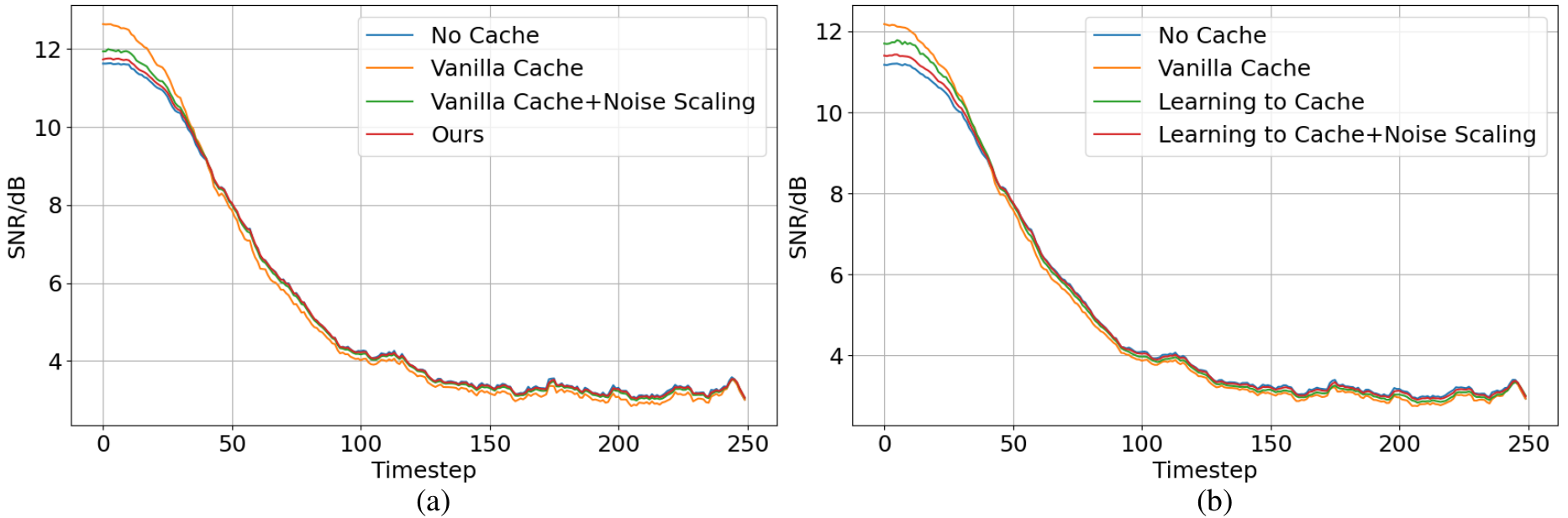}
    \captionsetup{font={normalsize,stretch=1.25},justification=raggedright}
    \caption{(a) Simply scaling the noise helps vanilla cache align with No-cache. Our method does a better job. (b) L2C also shows a tendency to align No-cache on SNR.}
    \label{fig: analysis}
\end{figure}

\begin{figure}[tp]
    \centering
    \includegraphics[width=0.95 \linewidth]{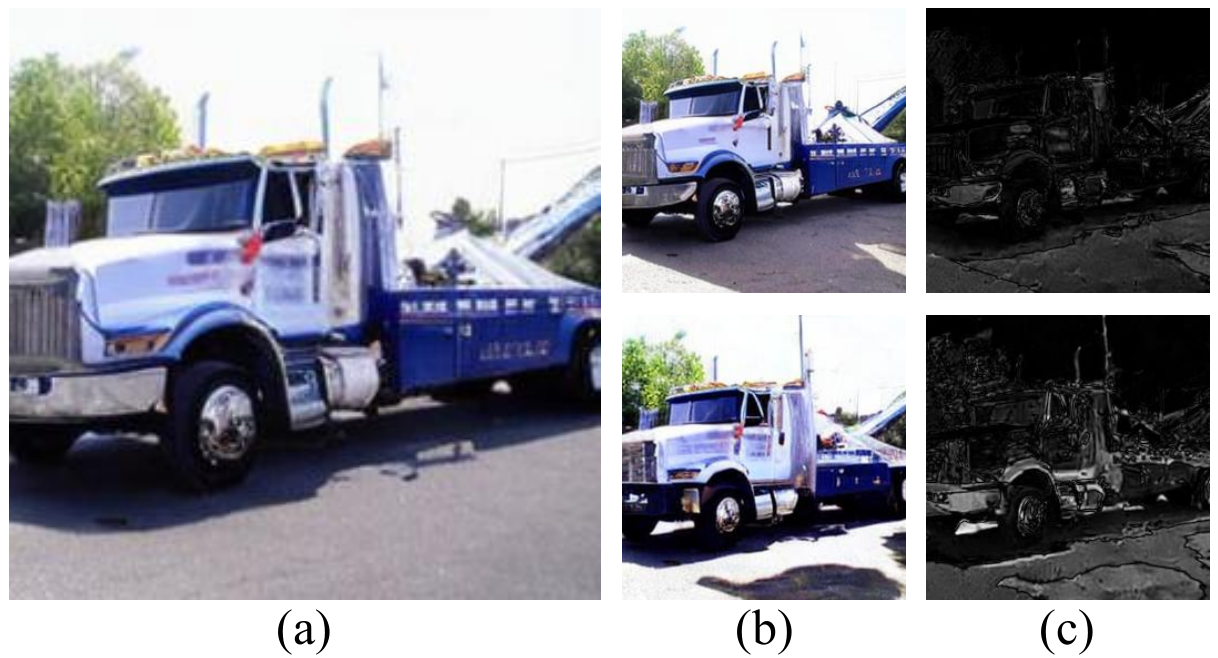}
    \captionsetup{font={normalsize,stretch=1.25},justification=raggedright}
    \caption{Feature and error visualization. (a) Ground truth. (b) Samples generated by ours (top) and vanilla cache (bottom). (c) Feature error between Ground truth and ours and vanilla cache at step 20. Our method shows lower error.}
    \label{fig: feature}
\end{figure}

\subsection{Ablation Study}

In this section, we conduct ablation studies on DiT/XL-2 with 50 DDPM steps to verify the contributions of core components and robustness of hyper-parameters.

\noindent \textbf{Component Ablation:} We first verify the contributions of the frequency-guided separated cache table and the stage-specific noise scaling as shown in Tab.~\ref{tab: ablation_to3}.

\noindent \textbf{Choice of Scaling Factor:} 
We illustrate the impact of different scaling factors in Tab.~\ref{tab: scaling_factor_ablation}. Note that 0.96 is a strong scaling compared to common noise scaling methods, (e.g., 0.989 for T=50 and 0.995 for T=100 in DiT-ES \cite{ningelucidating}). Since subsequent cache table searching compensates for over-correction of scale factors, our method demonstrates robustness across scale values. In practice, we determine scaling factors by searching within an empirical range.

\noindent \textbf{Influence of Threshold Function:} In Tab.~\ref{tab: threshold_ablation} and \ref{tab: threshold_value_ablation} we show the effects of different threshold function types and parameters($\delta(t)=a+b \cdot \frac{t}{T}, t\in[0,T)$). 
Beyond erroneous function forms (e.g., negative-linear violating commonly assumed degree of noise tolerance of the diffusion process), our method demonstrates robustness, with different settings consistently outperforming baseline approaches.

\noindent \textbf{Influence of $n$ in Algorithm.~\ref{algorithm}:} We conducted an ablation study on $n$ as shown in Tab.~\ref{tab: n_ablation}, increasing the number of samples almost makes no influence on the performance.

\subsection{Analysis}
To verify our initial conjecture, we show the SNR curve of uniform caching  in Fig.~\ref{fig: analysis}(a). 
Although not as good as our method, noise scaling for vanilla cache also shows a better SNR.
In addition, in Fig.~\ref{fig: analysis}(b) we find that L2C also outperforms vanilla cache, showing a tendency to align No-cache on SNR. At the same time, suitable noise reduction can further improve L2C.
We visualized the feature errors at the intermediate step versus the No-cache version in Fig.~\ref{fig: feature}.
The error of FEB-Cache and GT is only reflected in edges and the absolute error is marginal, while the vanilla cache has a large error in the structure, which deviates from the ideal sampling. More analysis can be found in Appendix.

\section{Conclusion}

In this paper, we point out the reason why DiT acceleration methods based on feature caching fail in quality: it greatly amplifies the exposure bias, and propose a method to achieve a balance between quality and speed with Epsilon Scaling . This is the first exploration in this area, bringing a new perspective to the community. We propose that separate cache table should be applied to the attention and MLP. Our method does not require heavy training and can be integrated with fast samplers. Experimental results show that compared with other acceleration strategies applied to DiT, this method is quite competitive and can significantly improve the speed while maintaining good quality.

\bibliography{aaai2026}

\clearpage

\noindent We organize the materials as follows:


\noindent 1.A sample of Cache Table and its performance

\noindent 2.Proof that exposure bias and variance are positively correlated

\noindent 3.More details and optimization of the signal in FEB-Cache

\noindent 4.More quantitative comparisons

\noindent 5.Visual comparison of exposure bias amplification

\noindent 6.Exposure Bias Reduction Works with DeepCache

\noindent 7.Derivation of {$\hat{{x}}_{t-1}$} and {$\hat{{x}}_{t-N}$}

\noindent 8.More Information about Separate Caching

\noindent 9.More Visualization Results

\noindent 10.Analysis of Acceleration for DiT

\noindent 11.Prompts for Images and Videos

\section{A sample of Cache Table and its performance}
\label{A sample of Cache Table}

\begin{figure}
    \centering
    \includegraphics[width=1 \linewidth]{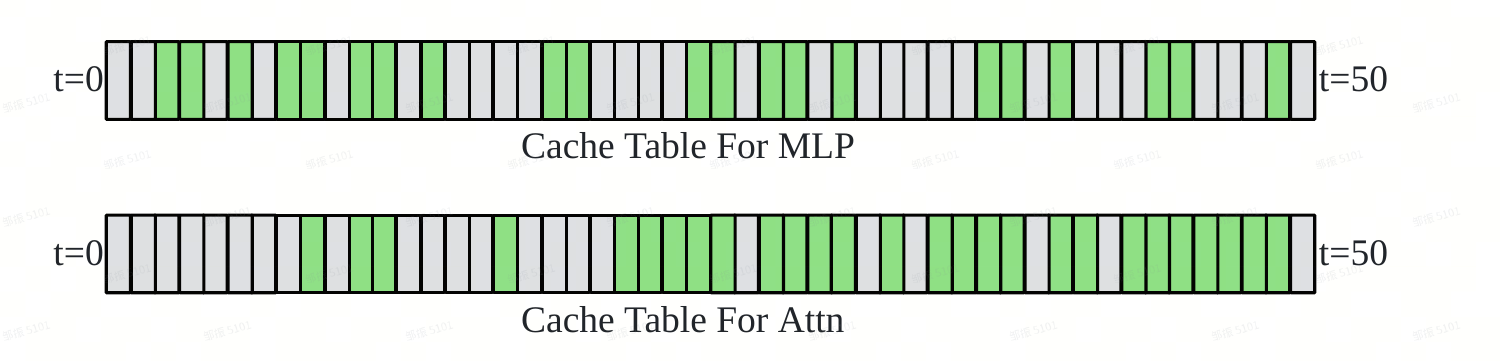}
    \caption{A sample of Cache Table for ImageNet 256 and 50 DDPM steps, MLP is cached more in high-noise phases, while Attn is cached more in low-noise phases.}
    \label{fig: cache_table}
\end{figure}

\begin{figure}
    \centering
    \includegraphics[width=1 \linewidth]{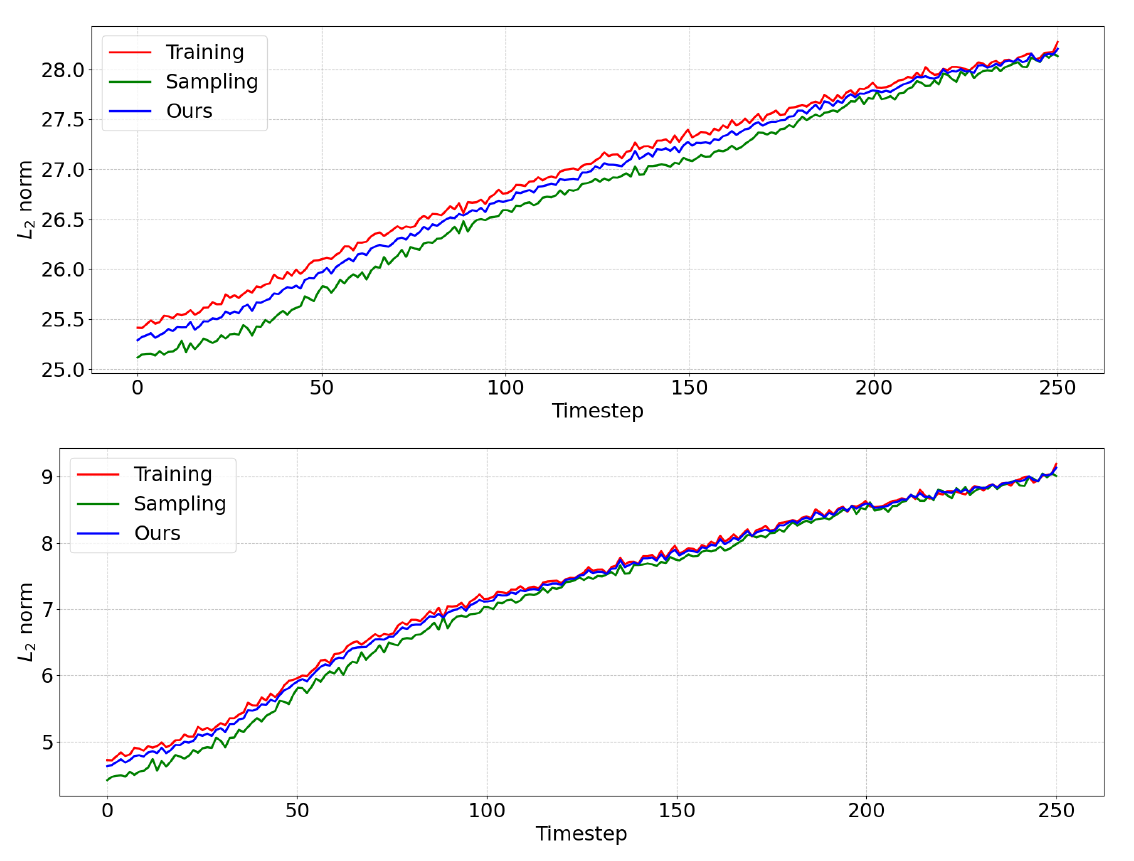}
    \caption{The effectiveness of our method in reducing exposure bias.}
    \label{fig: fig2_sup}
\end{figure}

In Fig.~\ref{fig: cache_table} we show a sample of Cache Table for ImageNet 256 and 50 DDPM steps.
Overall, MLP is cached more in high-noise phases, while Attn is cached more in low-noise phases.
Accordingly, Cache Table allows our method to move closer to the training process during sampling. We illustrate the effectiveness of our method with Fig.~\ref{fig: fig2_sup} corresponding to Fig.~2 in the main text.

\section{Exposure bias and variance are positively correlated.}
\label{proof that exposure bias and variance are positively correlated}

\textbf{Prerequisites for ddpm.} The forward process of the diffusion model is realized by gradually adding noise to the initial sample until pure Gaussian noise:
\begin{equation}
q({x}_t | {x}_{t-1}) = {\cal N} ({x}_t; \sqrt{1-{\beta}_t} {x}_{t-1}, {\beta}_t \pmb{I}), 
\end{equation}
where  $\{\beta_t\}^{T}_{t=1}$ Indicates the variance used at each step. Define $\alpha_t = 1 - \beta_t$ and $\bar{\alpha}_t = \prod \limits_{i=1}^t \alpha_i$, we can sample the $x_t$ of any step directly based on $x_0$:
\begin{equation}
    \label{eq: xt=x0+noise}
    {x}_t = \sqrt{\bar{\alpha}_t} {x}_0 +   \sqrt{1 - \bar{\alpha}_t} {\epsilon},
\end{equation}

In the reverse process of the diffusion model, the process of inferring $x_{t-1}$ from $x_t$ can be easily obtained by Bayes' theorem when $x_0$ is accessible:
\begin{equation}
    q({x}_{t-1} | {x}_{t}, {x}_{0}) = {\cal N} ({x}_{t-1}; {\tilde{\mu}}({x}_t, {x}_0), \tilde{\beta_t} \pmb{I})
    \label{eq: sample}
\end{equation}

\begin{equation}
    \label{eq: mu_x0}
    {\tilde{\mu}}({x}_t, {x}_0) = \frac{\sqrt{\alpha_{t}}(1-\bar{\alpha}_{t-1})}{1-\bar{\alpha}_{t}} {x}_{t} + \frac{\sqrt{\bar{\alpha}_{t-1}} \beta_{t}}{1-\bar{\alpha}_{t}} {x}_{0} 
\end{equation}
\begin{equation}
    \tilde{\beta_{t}} = \frac{1-\bar{\alpha}_{t-1}}{1-\bar{\alpha}_{t}}\beta_{t},
\end{equation}
where ${\tilde{\mu}}({x}_t, {x}_0)$ and $\tilde{\beta_{t}}$ indicate the mean and variance of $x_{t-1}$.
But $x_0$ is inaccessible during the inference stage and needs to be predicted by a network as $\hat{x}_0(x_t,t)$ :

\begin{align}
\hat{\mu}(x_t,t) & = \frac{\sqrt{\bar{\alpha}_{t-1}} \beta_{t}}{1-\bar{\alpha}_{t}} \hat{x}_0(x_t,t) + \frac{\sqrt{\alpha_{t}}(1-\bar{\alpha}_{t-1})}{1-\bar{\alpha}_{t}} {x}_{t} \label{eq: mu_theta} 
\end{align}

Exposure bias in diffusion models comes from the input mismatch between the training and sampling stages. During the sampling period, the model can't access $x_0$, leading to prediction errors. Suppose exposure bias occurs since step $t-1$ and the model's prediction $x_0$ in $t-1$ step marked as $\hat{x}_{0}(x_{t-1},t-1)$ can be modeled as a Gaussian distribution:

\begin{figure}
    \centering
    \includegraphics[width=1 \linewidth]{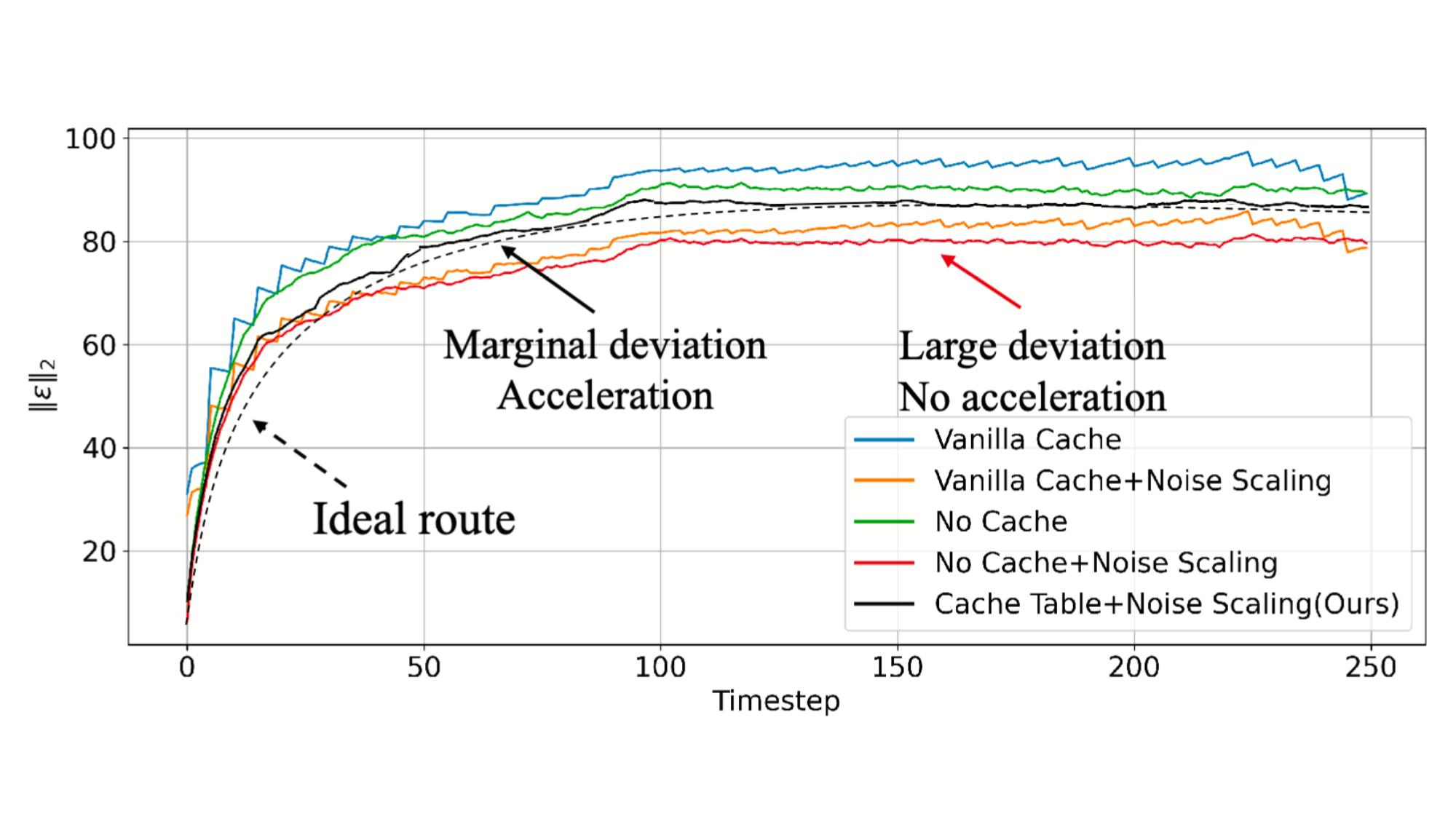}
    \caption{Noise Scaling exhibits a large deviation and no acceleration without Cache Table. Our method achieves marginal deviation and acceleration with Cache Table+Noise Scaling.}
    \label{fig: aaai1}
\end{figure}

\begin{equation}
        \hat{{x}}_{0}(x_{t-1},t-1) = {x}_{0} + e_{t-1} {\epsilon}_0 \quad ({\epsilon}_0 \sim {\cal N} ({0}, {I}))  
        \label{predict_error}
\end{equation}
where $e_{t-1} {\epsilon}_0$ refers to prediction error in sampling stage. 
Exposure bias comes from the inconsistency between training and inference in diffusion models. 
\begin{footnotesize}
\begin{equation}
\begin{aligned}
    \hat{x}_{t-1} &= \sqrt{\bar{\alpha}_{t-1}}x_0 + \sqrt{1 - \bar{\alpha}_{t-1} + \underbrace{\left(\frac{\sqrt{\bar{\alpha}_{t-1}}\beta_t}{1-\bar{\alpha}_t}e_t\right)^2}_{\text{Exposure Bias Term}}} \epsilon_1 ,
\end{aligned}
\end{equation}
\end{footnotesize}
where $Var(\hat{{x}}_{t-1}) = 1-\bar{\alpha}_{t-1} + (\frac{\sqrt{\bar{\alpha}_{t-1}} \beta_{t}}{1-\bar{\alpha}_{t}} e_{t} )^2 $ (See final two pages). $(\frac{\sqrt{\bar{\alpha}_{t-1}} \beta_{t}}{1-\bar{\alpha}_{t}} e_{t} )^2$ is only related with the noise schedule
and $e_t$, the variance of prediction error. When the variance increases, exposure bias rises accordingly cause the noise schedule is fixed. In the main body, we show the variance under $\rho$ coefficient. In final page, we prove the explosive growth for exposure bias when $p \rightarrow 1$.

\begin{figure}[htbp!]
    \centering
    \includegraphics[width=0.95\linewidth]{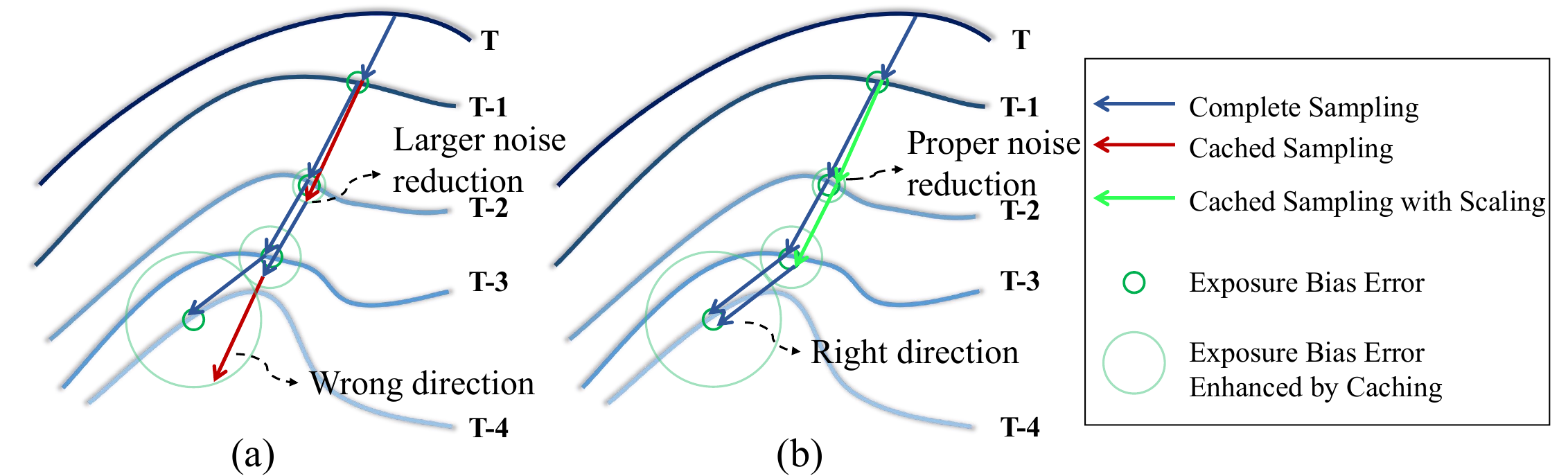}
    \caption{ Standard cache (a) predicts larger noise (amplification of exposure bias) and does not perceive the change of denoising direction.  Our method (b) scales the noise to keep the exposure bias error within an acceptable range. Cache Table senses direction changes and corrects.}
    \label{fig:aaaishiyi}
\end{figure}

\section{More details and optimization of the signal in FEB-Cache}
\label{More details and optimization of the signal in FEB-Cache}
Fig.~\ref{fig: aaai1} shows the impact of our optimized approach compared to simply caching features or noise scaling. Our method achieves marginal deviation and acceleration.
In Fig.~\ref{fig:aaaishiyi} we show an intuitive understanding of our method.

\section{More quantitative comparisons}
\label{More quantitative comparisons}
We show more quantitative comparisons on some other models and settings. We show the comparison with TeaCache (a superior approach to using feature caching) on FLUX (1k images) and Cosmos (50 videos) following TeaCache's setting is shown in Tab.~\ref{tab: main_table_rebuttal}. We achieved competitive results on PSNR, SSIM and LPIPS.
\begin{table}[tbh]
    \centering
    \caption{Comparison with Teacache.}
    \resizebox{0.49 \textwidth}{!}{
    \begin{tabular}{c|cccc}
        \hline
         Methods &  Latency(s) &  PSNR $\uparrow$ & SSIM $\uparrow$& LPIPS $\downarrow$ 
         \\
         \hline
        FLUX  &17.93&-&-&-
        \\
        TeaCache  &10.53&23.76&0.8560&0.2017
        \\
        Ours  & 10.46&\textbf{25.14}&\textbf{0.8741}&\textbf{0.1930}
        \\
        \hline
         \hline
        Cosmos-T2V  &409.60&-&-&-
        \\
        TeaCache  &301.74 &24.69&\textbf{0.8101}&0.1705
        \\
        Ours  & 299.13&\textbf{24.81}&0.8063&\textbf{0.1678}
        \\
        \hline
    \end{tabular}}
   \label{tab: main_table_rebuttal}
\end{table}

For separate caching on Attn and MLP, we show the result of the separate and unified caching in Tab.~\ref{tab: separate_caching} with the same setting as {Tab.2} in the main body but step=10.

\begin{table}[tbh]
    \centering
    \caption{Ablation on separate Self-Attn and MLP caching.}
    \resizebox{0.45 \textwidth}{!}{
    \begin{tabular}{c|cccccc}
        
        \hline
         &  Latency(s)  & FID $\downarrow$ &  sFID $\downarrow$ & IS $\uparrow$ & Precision$\uparrow$ & Recall$\uparrow$\\
        \hline
        DDIM(T=10)&  2.97  & 12.17 &  11.08 & 160.27 & 0.67 & 0.53 
        \\
        \hline
        DDIM(T=8)&  2.33  & 23.11 &  20.70 & 113.27 & 0.54 & 0.51
        \\
        \hline
        Uniform caching  & 2.31 & 18.19  &  16.81 & 127.75 & 0.61 & \textbf{0.53}
        \\
        \hline
        Separate caching & 2.27  & \textbf{16.67 } & \textbf{16.10} & \textbf{149.01} & \textbf{0.66} & 0.54
        \\
        \hline
    \end{tabular}}
    \label{tab: separate_caching}
\end{table}

\section{Visual comparison of exposure bias amplification}
\label{Visual comparison of exposure bias amplification}

\textbf{Objective:} Prove that even if adjacent error distributions are not identical but correlated, exposure bias accumulation still exceeds linear growth.

\subsection{Extended Assumptions}
Assume errors in adjacent steps are correlated with coefficient $\rho$ ($0 \leq \rho \leq 1$). The error at step $t-k$ can be expressed as:
$$
\epsilon_{t-k} = \rho \epsilon_{t-k+1} + \sqrt{1-\rho^2}\eta_k \quad (\eta_k \sim \mathcal{N}(0,I))
$$

\subsection{Variance Calculation}
For $N$ cached steps, the total variance of accumulated error is:
$$
\text{Var}\left(\sum_{k=1}^N \epsilon_{t-k}\right) = N + 2\sum_{d=1}^{N-1}(N-d)\rho^d
$$
where the covariance term dominates when $\rho \to 1$.

\subsection{Key Insight}
When $\rho$ is close to 1, variance grows super-linearly with $N$. Even for moderate $\rho$, the growth rate exceeds linear accumulation.

\subsection{Numerical Example: $\rho=0.8$, $N=5$}
\textit{This condition is pretty loose, and it can be seen from Figure 6 (b) of the mainbody that even without caching, the feature similarity of adjacent steps is high.}

\subsection{Calculation Details}
\begin{align*}
\text{Covariance Terms:} & \\
d=1: & \quad (5-1) \times 0.8^1 = 3.2 \\
d=2: & \quad (5-2) \times 0.8^2 = 1.92 \\
d=3: & \quad (5-3) \times 0.8^3 = 1.024 \\
d=4: & \quad (5-4) \times 0.8^4 = 0.4096 \\
\text{Total Covariance:} & \quad 3.2 + 1.92 + 1.024 + 0.4096 = 6.5536 \\
\text{Total Variance:} & \quad 5 + 2 \times 6.5536 = 18.1072 \\
\end{align*}

\begin{table}[h]
\centering
\caption{Variance Accumulation Comparison}
\begin{tabular}{l|cc}
 & \textbf{No Caching} & \textbf{Caching ($\rho=0.8$)} \\
\hline
Variance & 5 & 18.1072 \\
Std. Dev. & 2.236 & 4.256 \\
\end{tabular}
\end{table}

\subsection{Amplification Effects}
\noindent Variance amplification ratio: $18.1072/5 \approx 3.62\times$

\noindent Standard deviation ratio: $4.256/2.236 \approx 1.90\times$

\begin{figure*}
    \centering
    \includegraphics[width=\linewidth]{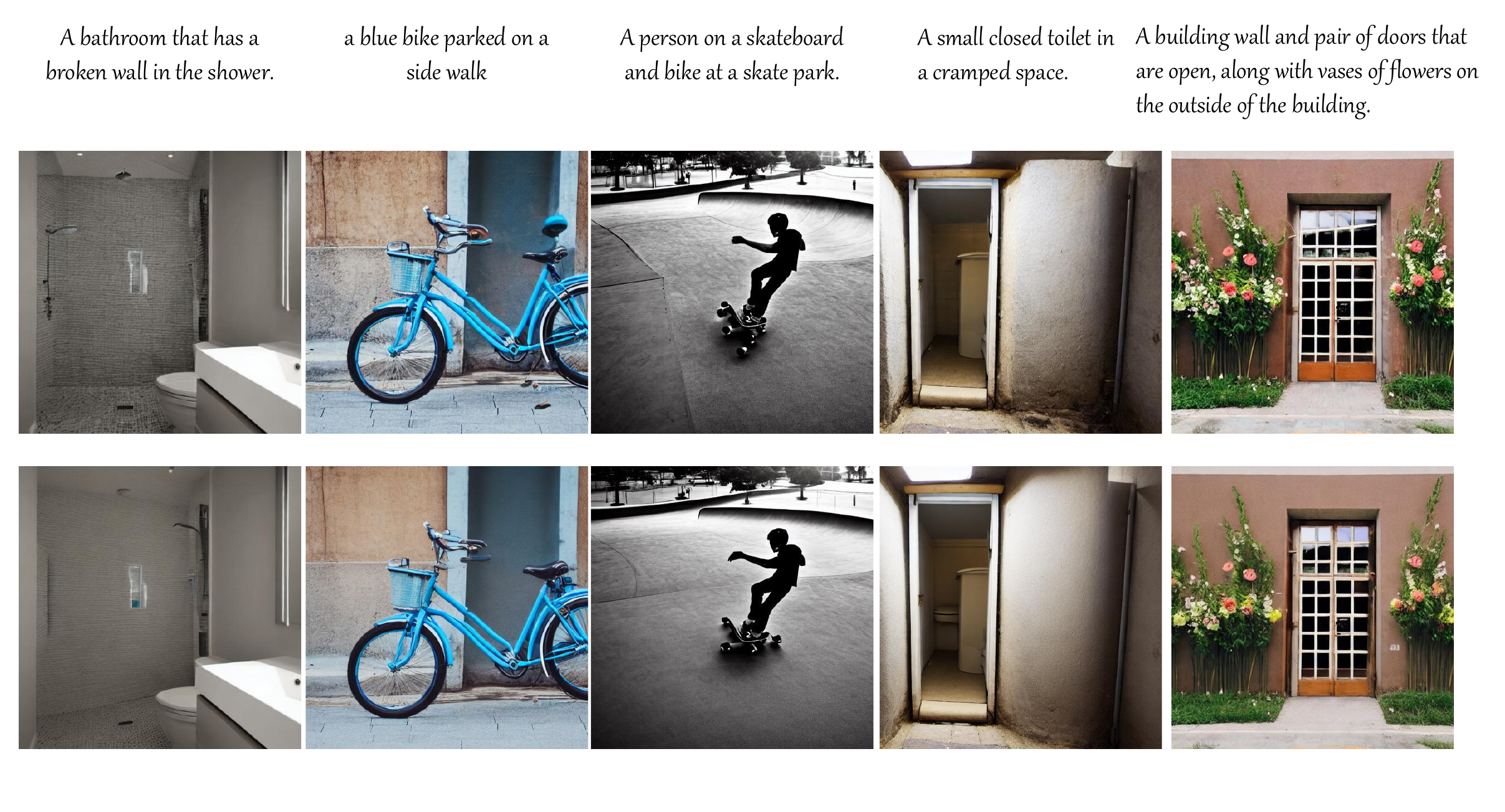}
    \vspace{-2em}
    \caption{Visualization of the generated images on COCO2017. Line 1 for DeepCache, line 2 for DeepCache+EBR.}
    \label{fig:cocoimg}
\end{figure*}

\noindent Even with $\rho=0.8$ (not perfect correlation), feature caching causes: 1. Variance growing $>3\times$ faster than linear accumulation and 2. Error magnitude nearly doubling

This demonstrates that the original core conclusion remains valid - exposure bias amplification through caching is significant even without strict identical distributions.

\begin{figure}
    \centering
    \includegraphics[width=1 \linewidth]{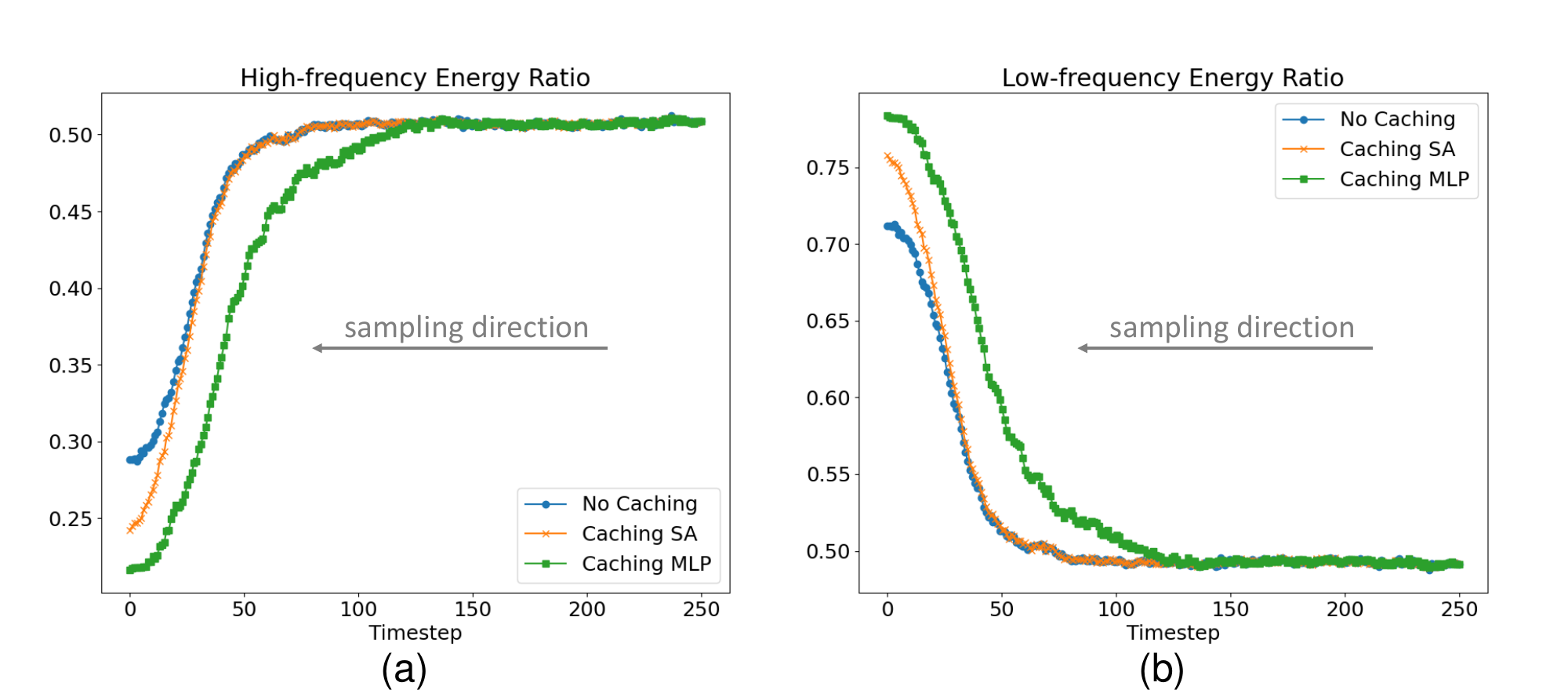}
    \caption{The proportion of frequency division energy during the diffusion process. MLP has a greater impact on frequency components and the stage is earlier.}
    \label{fig: supple_sa_mlp_intense}
\end{figure}

\section{Exposure Bias Reduction Works with DeepCache}
\label{Exposure Bias Reduction Works with DeepCache}
\begin{table}[t]
\centering
\caption{Exposure bias reduction works with DeepCache on Cifar-10. The experimental setup is the same as DeepCache.}
    \small
    \begin{tabular}{c | c c c c }
      \toprule
      \multicolumn{5}{c}{ FID   ON CIFAR-10 32 $\times$ 32 } \\
      \midrule
       &  N=2 &  N=3  &  N=5 &  N=10  \\
      \hline
       DeepCache         & 4.35 & 4.68 & 6.39 & 10.19 \\ \hline
       DeepCache + EBR  & 4.24 & 4.45 & 5.18 & 9.31  \\
       (Scaling Factor)&(0.996)&(0.995)&(0.995)&(0.994) \\
      
      %
      \bottomrule

    \end{tabular}

    \label{tbl:cifar}
\end{table}

Considering that exposure bias exists not only in DiT, we take DeepCache as an example to analyze the effectiveness of Exposure Bias Reduction (noise scaling) on non-transformer-based cache strategies. As shown in Table. \ref{tbl:cifar}, exposure bias reduction still applies in the context of DeepCache. We show visualization results on COCO2017 in \figurename~\ref{fig:cocoimg}. Images become relatively clear and conform to text semantics.

\section{Derivation of {$\hat{{x}}_{t-1}$} and {$\hat{{x}}_{t-N}$} }
\label{Full Derivation for Eq.4}


\textbf{Prerequisites for ddpm.} The forward process of the diffusion model is realized by gradually adding noise to the initial sample until pure Gaussian noise:
\begin{equation}
q({x}_t | {x}_{t-1}) = {\cal N} ({x}_t; \sqrt{1-{\beta}_t} {x}_{t-1}, {\beta}_t \pmb{I}), 
\end{equation}
where  $\{\beta_t\}^{T}_{t=1}$ Indicates the variance used at each step. Define $\alpha_t = 1 - \beta_t$ and $\bar{\alpha}_t = \prod \limits_{i=1}^t \alpha_i$, we can sample the $x_t$ of any step directly based on $x_0$:
\begin{equation}
    \label{eq: xt=x0+noise}
    {x}_t = \sqrt{\bar{\alpha}_t} {x}_0 +   \sqrt{1 - \bar{\alpha}_t} {\epsilon},
\end{equation}

In the reverse process of the diffusion model, the process of inferring $x_{t-1}$ from $x_t$ can be easily obtained by Bayes' theorem when $x_0$ is accessible:
\begin{equation}
    q({x}_{t-1} | {x}_{t}, {x}_{0}) = {\cal N} ({x}_{t-1}; {\tilde{\mu}}({x}_t, {x}_0), \tilde{\beta_t} \pmb{I})
    \label{eq: sample}
\end{equation}

\begin{equation}
    \label{eq: mu_x0}
    {\tilde{\mu}}({x}_t, {x}_0) = \frac{\sqrt{\alpha_{t}}(1-\bar{\alpha}_{t-1})}{1-\bar{\alpha}_{t}} {x}_{t} + \frac{\sqrt{\bar{\alpha}_{t-1}} \beta_{t}}{1-\bar{\alpha}_{t}} {x}_{0} 
\end{equation}
\begin{equation}
    \tilde{\beta_{t}} = \frac{1-\bar{\alpha}_{t-1}}{1-\bar{\alpha}_{t}}\beta_{t},
\end{equation}
where ${\tilde{\mu}}({x}_t, {x}_0)$ and $\tilde{\beta_{t}}$ indicate the mean and variance of $x_{t-1}$.
But $x_0$ is inaccessible during the inference stage and needs to be predicted by a network as $\hat{x}_0(x_t,t)$ :

\begin{figure}
    \centering
    \includegraphics[width=1 \linewidth]{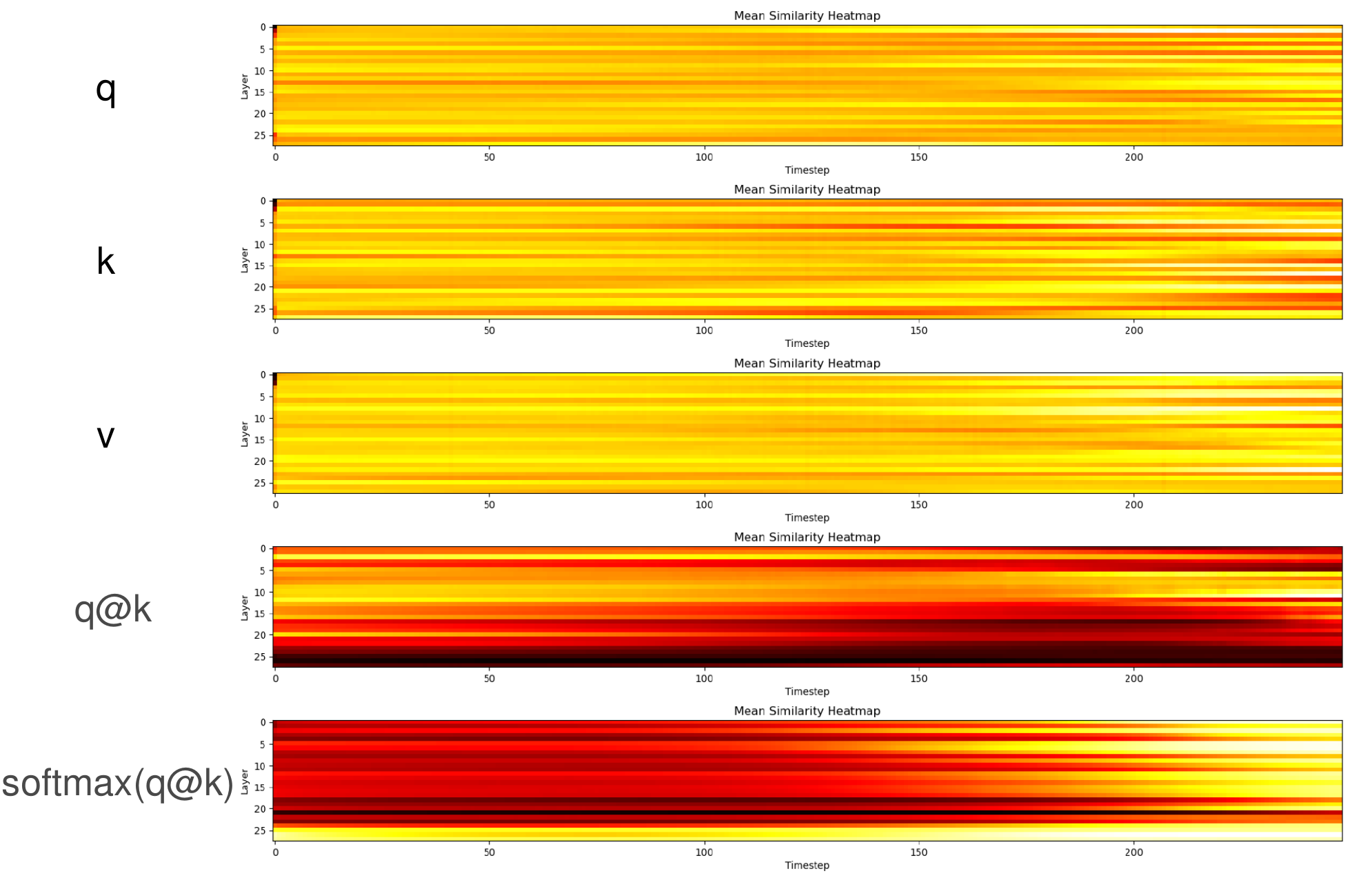}
    \caption{Similarity heat map for components in SA. The same settings are used in Figure 5 in the manuscript.}
    \label{fig: supple_similarity}
\end{figure}

\begin{align}
\hat{\mu}(x_t,t) & = \frac{\sqrt{\bar{\alpha}_{t-1}} \beta_{t}}{1-\bar{\alpha}_{t}} \hat{x}_0(x_t,t) + \frac{\sqrt{\alpha_{t}}(1-\bar{\alpha}_{t-1})}{1-\bar{\alpha}_{t}} {x}_{t} \label{eq: mu_theta} 
\end{align}

Exposure bias in diffusion models comes from the input mismatch between the training and sampling stages. During the sampling period, the model can't access $x_0$, leading to prediction errors. Suppose exposure bias occurs since step $t-1$ and the model's prediction $x_0$ in $t-1$ step marked as $\hat{x}_{0}(x_{t-1},t-1)$ can be modeled as a Gaussian distribution:

\begin{equation}
        \hat{{x}}_{0}(x_{t-1},t-1) = {x}_{0} + e_{t-1} {\epsilon}_0 \quad ({\epsilon}_0 \sim {\cal N} ({0}, {I}))  
        \label{predict_error}
\end{equation}
where $e_t {\epsilon}_0$ refers to prediction error in sampling stage. For clarity, we give the full proof of $\hat{{x}}_{t-N}$ 

\section{More Information about Separate Caching for Attention and MLP}
\label{More Information about Separate Caching for Attention and MLP}

\begin{figure}
    \centering
    \includegraphics[width=1 \linewidth]{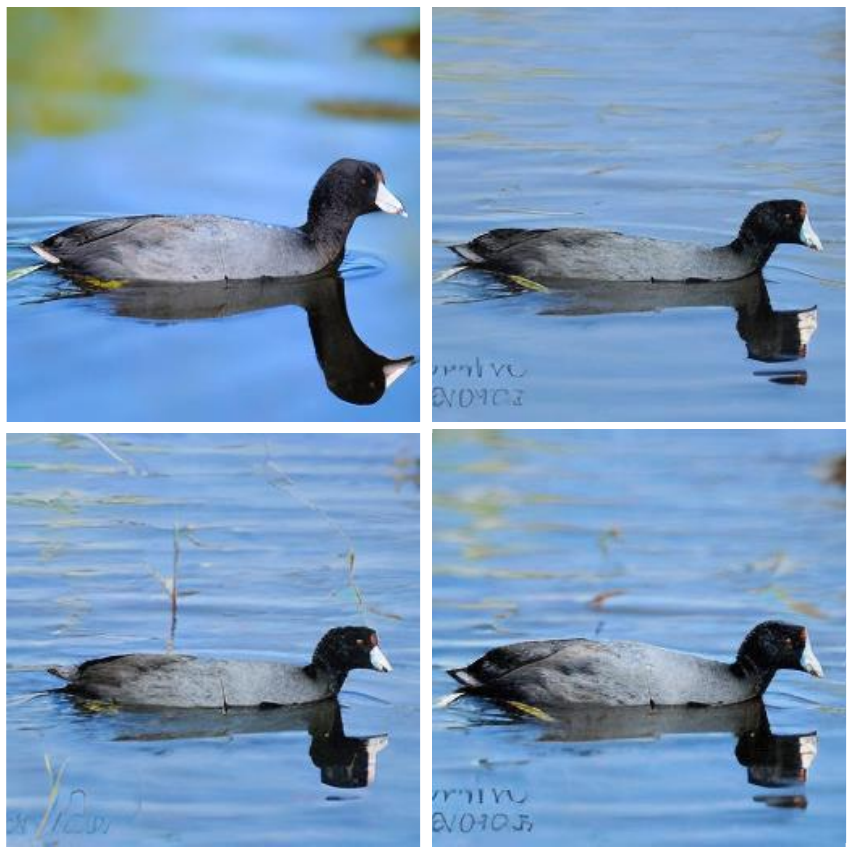}
    \caption{Four different caching strategies. Top left: caching MLP; top right: caching SA; lower left: caching both; lower right: no caching. The caching intervals are set to 5.}
    \label{fig: supple_sa_mlp}
\end{figure}

Caching MLP reduces the high-frequency components of the image and makes it too smooth. In \figurename~\ref{fig: supple_sa_mlp_intense}, we show the curve of the frequency-division energy intensity during the diffusion process on 10k samples. It's easy to find that caching MLP reduces more high-frequency signal ($29\% \rightarrow 21\%$) at an earlier stage (almost from 125 steps).
In \figurename~\ref{fig: supple_sa_mlp}, we illustrate the impact of four distinct caching strategies. Caching MLP (upper left) solely significantly obliterates image details, exerting a substantial negative effect on the generated outcomes, and is even inferior to caching both (lower left). Conversely, caching only SA (upper right) has minimal impact, with a negligible difference from the original image (lower right), and no discernible artifacts are introduced.

In addition to the separability of SA and MLP, considering the calculation process of SA, we further analyzed the similarities of its more detailed components.
\begin{align}
    &attn = q @ k.transpose(-2, -1) \nonumber\\
    &attn = attn.softmax(dim=-1) \nonumber\\
    &attn = self.attn\_drop(attn) \nonumber\\
    &x = attn @ v \nonumber
\end{align}

We further analyzed the similarity of its detailed components as shown in \figurename~\ref{fig: supple_similarity}. Unfortunately, although q, k, and v all show high similarity, the elements after nonlinear operations become less similar and less cacheable. This prevents us from further subdividing SA.

\begin{figure}
    \centering
    \includegraphics[width=0.95\linewidth]{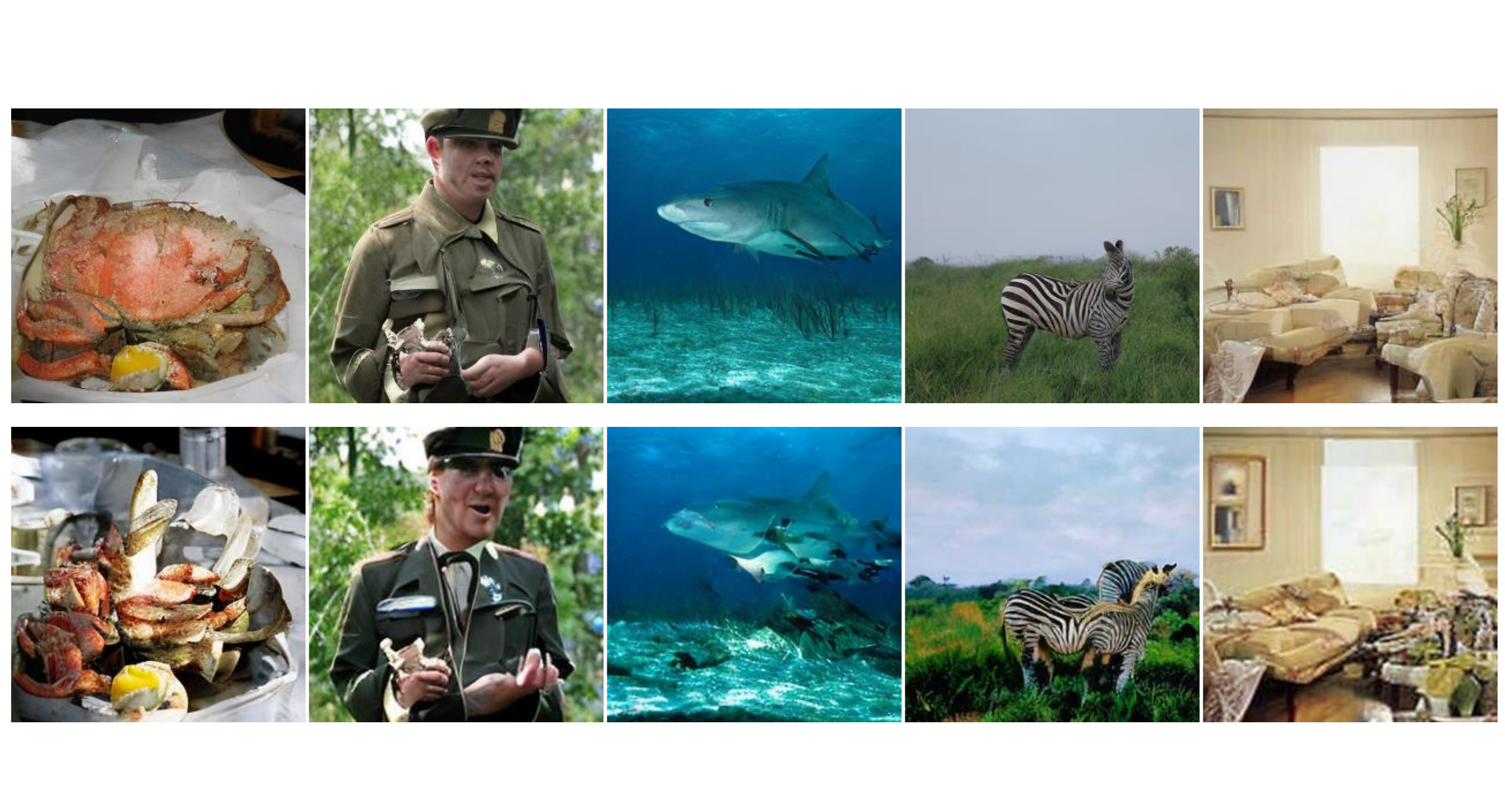}
    \caption{250 NFEs generation on ImageNet 256 $\times$ 256 with DiTFastAttn \cite{yuan2024ditfastattn}. It performs poorly in low-resolution acceleration.}
    \label{fig:ditfastattn}
\end{figure}

\section{More Visualization Results}
\label{More Visualization Results}
\begin{figure*}
    \centering
    \includegraphics[width=0.95 \linewidth]{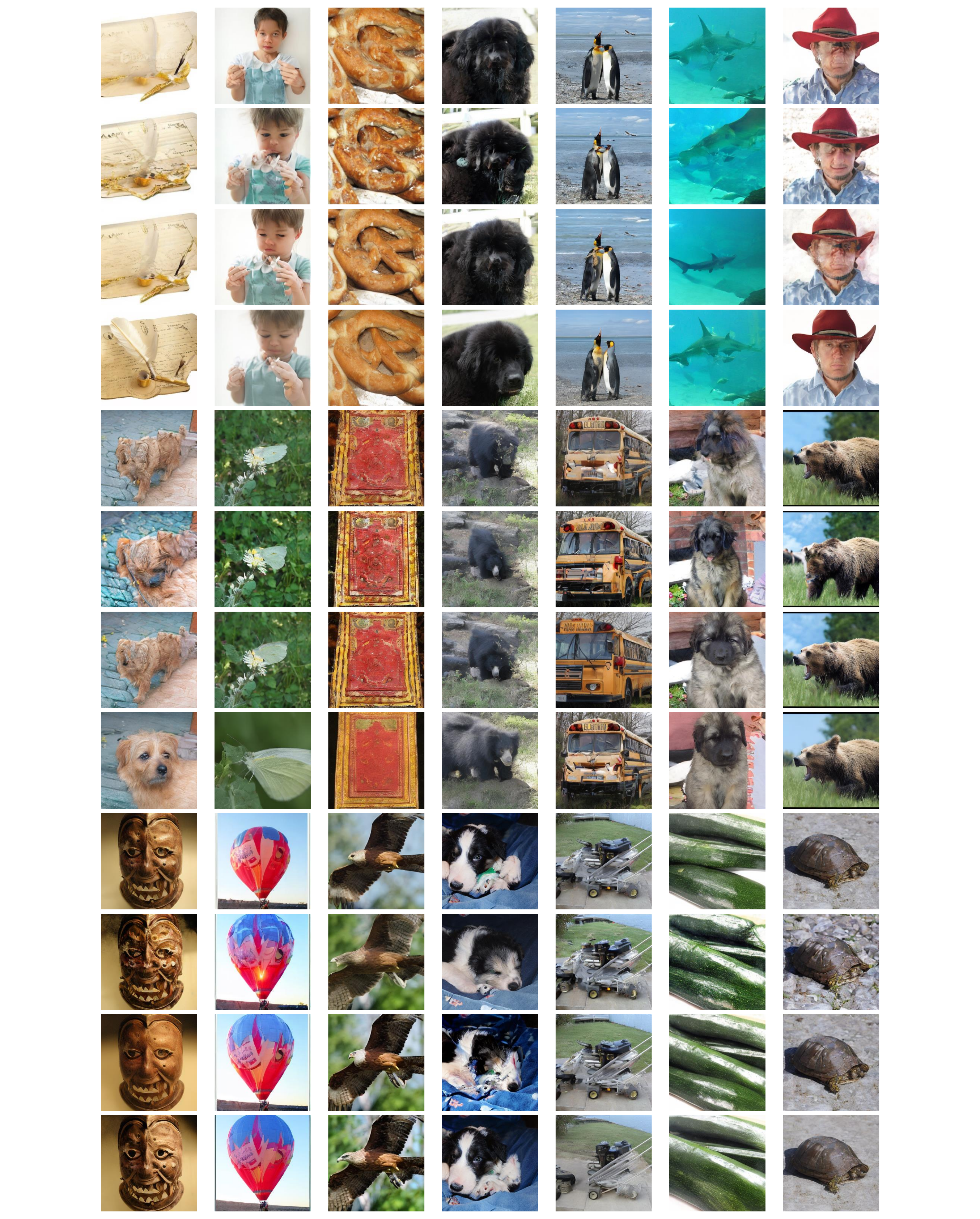}
    \caption{Samples on ImageNet 256 $\times$ 256 with 50NFEs (line 1), FORA (line 2, speedup ratio = 1.55$\times$), L2C (line3, speedup ratio = 1.30$\times$) and 50 NFEs+FEB-Cache (line4, speedup ratio = 1.53$\times$).}
    \label{fig: 50v}
    \vspace{-3em}
\end{figure*}

\begin{figure*}
    \centering
    \includegraphics[width=0.95 \linewidth]{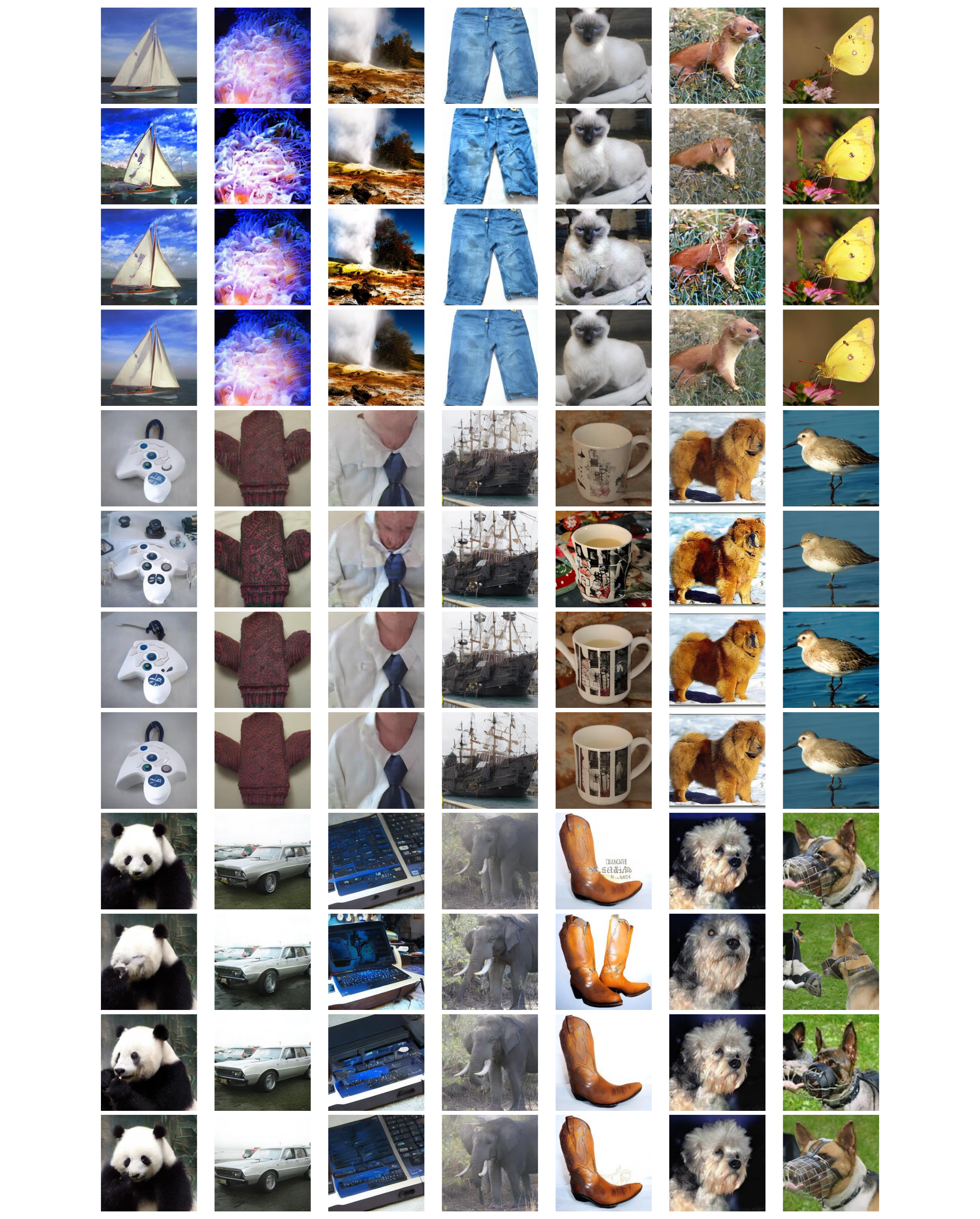}
    \caption{Samples on ImageNet 256 $\times$ 256 with 100NFEs (line 1), FORA (line 2, speedup ratio = 1.55$\times$), L2C (line 3, speedup ratio = 1.29$\times$), and 100 NFEs+FEB-Cache (line 4, speedup ratio = 1.52$\times$).}
    \label{fig: 100v}
    \vspace{-3em}
\end{figure*}

\begin{figure*}
    \centering
    \includegraphics[width=0.95 \linewidth]{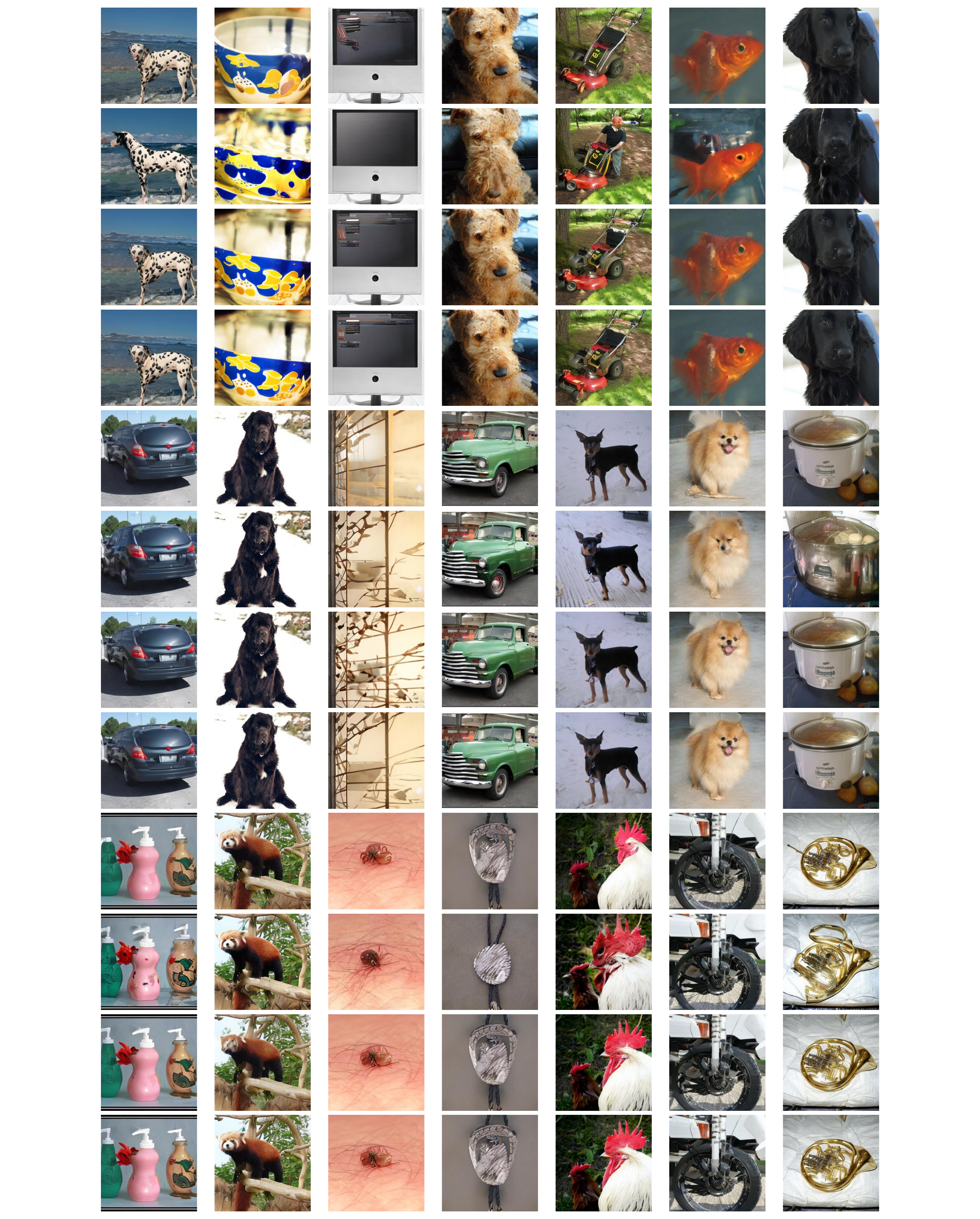}
    \caption{Samples on ImageNet 256 $\times$ 256 with 250NFEs (line 1), FORA (line 2, speedup ratio = 2.46$\times$), L2C (line 3, speedup ratio = 1.65$\times$), and 250 NFEs+FEB-Cache (line 4, speedup ratio = 2.48$\times$).}
    \label{fig: 250v}
    \vspace{-3em}
\end{figure*}

We show more visual comparisons on ImageNet in \figurename~\ref{fig: 50v}, \ref{fig: 100v}, and \ref{fig: 250v}. Comparisons with other cache-based methods (FORA, L2C) are also conducted. As shown in \figurename~\ref{fig:ditfastattn}, DiTFastAttn performs poorly in low-resolution (256 $\times$ 256) acceleration so we do not include it in  \figurename~\ref{fig: 50v}, \ref{fig: 100v}, and \ref{fig: 250v}.

\section{Analysis of Acceleration for DiT}
\label{Analysis of Acceleration for DiT}
DeepCache has achieved good results in the U-Net architecture. Unlike the U-Net-based diffusion model, DiT is isotropic and has no skip connections, which leads to strong inter-layer coupling \cite{chen2024delta}. Previous methods \cite{selvaraju2024fora,ma2024learning,yuan2024ditfastattn} have difficulty in achieving a high speedup ratio (e.g. 5$\times$) while ensuring generation quality. Our approach has gone a step further by starting from the impact of cache on the diffusion process, but further exploration is needed to achieve a perfect balance between high speedup ratio and quality.

\section{Prompts for Images and Videos}
\label{Prompts for Images and Videos}
Prompts for Fig. 8 in the mainbody are as follows:
\begin{itemize}
    \item A small cactus with a happy face in the Sahara desert.
    \item Deadpool,half body shot,In the morning mist.
    \item A classic motorcycle parked in a maple forest.
\end{itemize}

\begin{figure*}
    \centering
    \includegraphics[width=0.95 \linewidth]{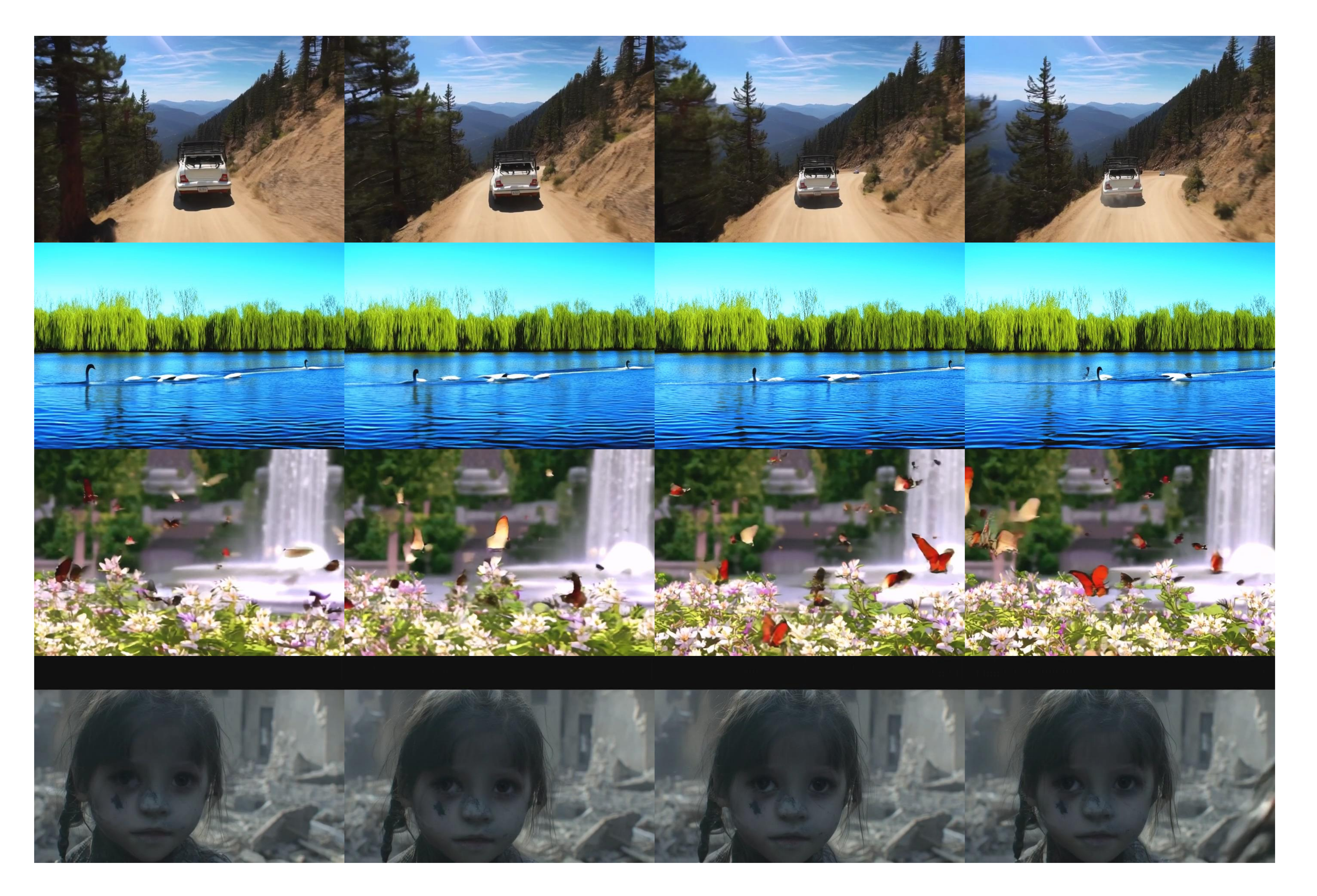}
    \caption{Samples 720 $\times$ 480 CogvideoX (speedup ration=1.45$\times$)}
    \label{fig: video}
    \vspace{-3em}
\end{figure*}

\noindent Prompts for Fig.~\ref{fig: video} are as follows:
\begin{itemize}

    \item The camera follows behind a white vintage SUV with a black roof rack as it speeds up a steep dirt road surrounded by pine trees on a steep mountain slope, dust kicks up from its tires, the sunlight shines on the SUV as it speeds along the dirt road, casting a warm glow over the scene. The dirt road curves gently into the distance, with no other cars or vehicles in sight. The trees on either side of the road are redwoods, with patches of greenery scattered throughout. The car is seen from the rear following the curve with ease, making it seem as if it is on a rugged drive through the rugged terrain. The dirt road itself is surrounded by steep hills and mountains, with a clear blue sky above with wispy clouds.
    \item On a brilliant sunny day, the lakeshore is lined with an array of willow trees, their slender branches swaying gently in the soft breeze. The tranquil surface of the lake reflects the clear blue sky, while several elegant swans glide gracefully through the still water, leaving behind delicate ripples that disturb the mirror-like quality of the lake. The scene is one of serene beauty, with the willows' greenery providing a picturesque frame for the peaceful avian visitors.
    \item  A garden comes to life as a kaleidoscope of butterflies flutters amidst the blossoms, their delicate wings casting shadows on the petals below.  In the background, a grand fountain cascades water with a gentle splendor, its rhythmic sound providing a soothing backdrop.  Beneath the cool shade of a mature tree, a solitary wooden chair invites solitude and reflection, its smooth surface worn by the touch of countless visitors.
    \item In the haunting backdrop of a war-torn city, where ruins and crumbled walls tell a story of devastation, a poignant close-up frames a young girl. Her face is smudged with ash, a silent testament to the chaos around her. Her eyes glistening with a mix of sorrow and resilience, capturing the raw emotion of a world that has lost its innocence to the ravages of conflict.
    
\end{itemize}

\clearpage
\Large
\raggedright
\textbf{Proof of $\pmb{\hat{{x}}}_{\pmb{t-1}}$ }
\normalsize
\begin{strip}
Plug Eq. \ref{eq: mu_theta}, Eq. \ref{eq: xt=x0+noise} and $\hat{{x}}_{0}(x_{t-1},t-1)$ to Eq. \ref{eq: sample}:
\begin{align}
    \hat{x}_{t-1} &= \hat{\mu}(x_{t},t) + \sqrt{\tilde{\beta}_{t}}\epsilon_2 \nonumber \\ \nonumber \\
    &= \frac{\sqrt{\bar{\alpha}_{t-1}} \beta_{t}}{1-\bar{\alpha}_{t}} \hat{{x}}_{0}(x_{t-1},t-1) + \frac{\sqrt{\alpha_{t}}(1-\bar{\alpha}_{t-1})}{1-\bar{\alpha}_{t}} {x}_{t} + \sqrt{\tilde{\beta}_{t}}\epsilon_2 \nonumber \\ \nonumber \\
    &=\frac{\sqrt{\bar{\alpha}_{t-1}} \beta_{t}}{1-\bar{\alpha}_{t}} ({x}_{0} + e_{t-1} {\epsilon}_0) +\frac{\sqrt{\alpha_{t}}(1-\bar{\alpha}_{t-1})}{1-\bar{\alpha}_{t}} (\sqrt{\bar{\alpha}_{t}}x_0+\sqrt{1-\bar{\alpha}_{t}}\epsilon) +\sqrt{\tilde{\beta}_{t}}\epsilon_2 \nonumber \\
    &= \frac{\sqrt{\bar{\alpha}_{t-1}} \beta_{t} + \sqrt{\alpha_{t}}(1-\bar{\alpha}_{t-1}) \sqrt{\bar{\alpha}_{t}} }{1-\bar{\alpha}_{t}}  {x}_{0} + \frac{\sqrt{\bar{\alpha}_{t-1}} \beta_{t}}{1-\bar{\alpha}_{t}} e_{t}{\epsilon}_0  + \sqrt{ \tilde{\beta}_{t}} {\epsilon}_2 + \frac{\sqrt{\alpha_{t}}(1-\bar{\alpha}_{t-1})}{1-\bar{\alpha}_{t}} \sqrt{1 -\bar{\alpha}_{t}} {\epsilon} \nonumber \\ \nonumber \\
    &= \frac{\sqrt{\bar{\alpha}_{t-1}} (1-\alpha_{t}) + \alpha_{t}(1-\bar{\alpha}_{t-1}) \sqrt{\bar{\alpha}_{t-1}} }{1-\bar{\alpha}_{t}}  {x}_{0} +\frac{\sqrt{\bar{\alpha}_{t-1}} \beta_{t}}{1-\bar{\alpha}_{t}} e_{t-1}{\epsilon}_0  + \sqrt{ \tilde{\beta}_{t}} {\epsilon}_2  + \frac{\sqrt{\alpha_{t}}(1-\bar{\alpha}_{t-1})}{1-\bar{\alpha}_{t}} \sqrt{1 -\bar{\alpha}_{t}} {\epsilon}  \nonumber \\
    &= \sqrt{\bar{\alpha}_{t-1}} x_0 +\frac{\sqrt{\bar{\alpha}_{t-1}} \beta_{t}}{1-\bar{\alpha}_{t}} e_{t}{\epsilon}_0  + \sqrt{ \tilde{\beta}_{t}} {\epsilon}_2 +  \frac{\sqrt{\alpha_{t}}(1-\bar{\alpha}_{t-1})}{1-\bar{\alpha}_{t}} \sqrt{1 -\bar{\alpha}_{t}} {\epsilon} 
    \label{t-1 proof}
\end{align}
where $\epsilon, \epsilon_0 $ and $\epsilon_2$ are uncorrelated. Let's focus on the variance 
\begin{align}
    Var(\hat{x}_{t-1}) &= (\frac{\sqrt{\bar{\alpha}_{t-1}} \beta_{t}}{1-\bar{\alpha}_{t}} e_{t} )^2 + (\frac{\sqrt{\alpha_{t}}(1-\bar{\alpha}_{t-1})}{1-\bar{\alpha}_{t}} \sqrt{1 -\bar{\alpha}_{t}} )^2 + \tilde{\beta}_{t} 
\end{align}
where $(\frac{\sqrt{\alpha_{t}}(1-\bar{\alpha}_{t-1})}{1-\bar{\alpha}_{t}} \sqrt{1 -\bar{\alpha}_{t}} )^2 + \tilde{\beta}_{t} $ can be easily simplified to $1-\bar{\alpha}_{t-1}$. So we get the variance of $\hat{x}_{t-1}$:
\begin{equation}
    Var(\hat{x}_{t-1}) =  ( \frac{\sqrt{\bar{\alpha}_{t-1}} \beta_{t}}{1-\bar{\alpha}_{t}} e_{t} )^2 + 1-\bar{\alpha}_{t-1}
    \label{total_1_eq}
\end{equation}

\end{strip}
\clearpage

\clearpage
\Large
\raggedright
\textbf{Proof of $\pmb{\hat{{x}}}_{\pmb{t-N}}$ }
\normalsize
\begin{strip}
Since the same noise distribution is superimposed on $N$ steps of continuous caching, we can extend Eq. \ref{predict_error} to $N$ steps $\hat{{x}}_{0}(x_{t-N},t-N) = {x}_{0} + N \cdot e_{t-1} {\epsilon}_0$. Plug Eq. \ref{eq: mu_theta}, Eq. \ref{eq: xt=x0+noise} and $\hat{{x}}_{0}(x_{t-N},t-N)$ to Eq. \ref{eq: sample}:
\begin{align}
    \hat{x}_{t-N} &= \hat{\mu}(x_{t-N+1},t-N+1) + \sqrt{\tilde{\beta}_{t-N+1}}\epsilon_2 \nonumber \\ \nonumber \\
    &= \frac{\sqrt{\bar{\alpha}_{t-N}} \beta_{t-N+1}}{1-\bar{\alpha}_{t-N+1}} \hat{{x}}_{0}(x_{t-N},t-N) + \frac{\sqrt{\alpha_{t-N+1}}(1-\bar{\alpha}_{t-N})}{1-\bar{\alpha}_{t-N+1}} {x}_{t-N+1} + \sqrt{\tilde{\beta}_{t-N+1}}\epsilon_2 \nonumber \\ \nonumber \\
    &=\frac{\sqrt{\bar{\alpha}_{t-N}} \beta_{t-N+1}}{1-\bar{\alpha}_{t-N+1}} ({x}_{0} + N \cdot e_{t-1} {\epsilon}_0) +\frac{\sqrt{\alpha_{t-N+1}}(1-\bar{\alpha}_{t-N})}{1-\bar{\alpha}_{t-N+1}} (\sqrt{\bar{\alpha}_{t-N+1}}x_0+\sqrt{1-\bar{\alpha}_{t-N+1}}\epsilon) +\sqrt{\tilde{\beta}_{t-N+1}}\epsilon_2 \nonumber \\ \nonumber \\
    &= \frac{\sqrt{\bar{\alpha}_{t-N}} \beta_{t-N+1} + \sqrt{\alpha_{t-N+1}}(1-\bar{\alpha}_{t-N}) \sqrt{\bar{\alpha}_{t-N+1}} }{1-\bar{\alpha}_{t-N+1}}  {x}_{0} + N \cdot\frac{\sqrt{\bar{\alpha}_{t-N}} \beta_{t-N+1}}{1-\bar{\alpha}_{t-N+1}} e_{t-N+1}{\epsilon}_0  + \sqrt{ \tilde{\beta}_{t-N+1}} {\epsilon}_2 \nonumber \\
    & + \frac{\sqrt{\alpha_{t-N+1}}(1-\bar{\alpha}_{t-N})}{1-\bar{\alpha}_{t-N+1}} \sqrt{1 -\bar{\alpha}_{t-N+1}} {\epsilon} \nonumber \\ \nonumber \\
    &= \frac{\sqrt{\bar{\alpha}_{t-N}} (1-\alpha_{t-N+1}) + \alpha_{t-N+1}(1-\bar{\alpha}_{t-N}) \sqrt{\bar{\alpha}_{t-N}} }{1-\bar{\alpha}_{t-N+1}}  {x}_{0} + N \cdot\frac{\sqrt{\bar{\alpha}_{t-N}} \beta_{t-N+1}}{1-\bar{\alpha}_{t-N+1}} e_{t-N+1}{\epsilon}_0  + \sqrt{ \tilde{\beta}_{t-N+1}} {\epsilon}_2 \nonumber \\
    & + \frac{\sqrt{\alpha_{t-N+1}}(1-\bar{\alpha}_{t-N})}{1-\bar{\alpha}_{t-N+1}} \sqrt{1 -\bar{\alpha}_{t-N+1}} {\epsilon} \nonumber \\ \nonumber \\
    &= \sqrt{\bar{\alpha}_{t-N}} x_0 + N \cdot\frac{\sqrt{\bar{\alpha}_{t-N}} \beta_{t-N+1}}{1-\bar{\alpha}_{t-N+1}} e_{t-N+1}{\epsilon}_0  + \sqrt{ \tilde{\beta}_{t-N+1}} {\epsilon}_2 +  \frac{\sqrt{\alpha_{t-N+1}}(1-\bar{\alpha}_{t-N})}{1-\bar{\alpha}_{t-N+1}} \sqrt{1 -\bar{\alpha}_{t-N+1}} {\epsilon} 
    \label{t-N proof}
\end{align}
where $\epsilon, \epsilon_0 $ and $\epsilon_2$ are uncorrelated. Let's focus on the variance 
\begin{align}
    Var(\hat{x}_{t-N}) &= (N \cdot \frac{\sqrt{\bar{\alpha}_{t-N}} \beta_{t-N+1}}{1-\bar{\alpha}_{t-N+1}} e_{t-N+1} )^2 + (\frac{\sqrt{\alpha_{t-N+1}}(1-\bar{\alpha}_{t-N})}{1-\bar{\alpha}_{t-N+1}} \sqrt{1 -\bar{\alpha}_{t-N+1}} )^2 + \tilde{\beta}_{t-N+1} 
\end{align}
where $(\frac{\sqrt{\alpha_{t-N+1}}(1-\bar{\alpha}_{t-N})}{1-\bar{\alpha}_{t-N+1}} \sqrt{1 -\bar{\alpha}_{t-N+1}} )^2 + \tilde{\beta}_{t-N+1} $ can be easily simplified to $1-\bar{\alpha}_{t-N}$. So we get the variance of $\hat{x}_{t-N}$:
\begin{equation}
    Var(\hat{x}_{t-N}) = N^2 ( \frac{\sqrt{\bar{\alpha}_{t-N}} \beta_{t-N+1}}{1-\bar{\alpha}_{t-N+1}} e_{t-N+1} )^2 + 1-\bar{\alpha}_{t-N}
    \label{total_eq}
\end{equation}
Now that we have determined the mean and variance of $\hat{x}_{t-N}$.
\end{strip}
\clearpage


\end{document}